\documentclass{article} 
\usepackage{iclr2023_conference,times}

\usepackage{amsmath,amsfonts,bm}

\def\eqref#1{equation~\ref{#1}}

\def\1{\bm{1}}

\def\ra{{\textnormal{a}}}

\def\rx{{\textnormal{x}}}

\def\rva{{\mathbf{a}}}

\def\erva{{\textnormal{a}}}

\def\ervx{{\textnormal{x}}}

\def\rmA{{\mathbf{A}}}

\def\vmu{{\bm{\mu}}}
\def\vtheta{{\bm{\theta}}}
\def\va{{\bm{a}}}

\def\ve{{\bm{e}}}

\def\vx{{\bm{x}}}

\def\eva{{a}}

\def\mA{{\bm{A}}}

\def\mH{{\bm{H}}}
\def\mI{{\bm{I}}}
\def\mJ{{\bm{J}}}

\def\mX{{\bm{X}}}

\def\mSigma{{\bm{\Sigma}}}

\DeclareMathAlphabet{\mathsfit}{\encodingdefault}{\sfdefault}{m}{sl}
\SetMathAlphabet{\mathsfit}{bold}{\encodingdefault}{\sfdefault}{bx}{n}
\newcommand{\tens}[1]{\bm{\mathsfit{#1}}}
\def\tA{{\tens{A}}}

\def\tX{{\tens{X}}}

\def\gG{{\mathcal{G}}}

\def\sA{{\mathbb{A}}}
\def\sB{{\mathbb{B}}}

\def\sS{{\mathbb{S}}}

\def\emA{{A}}

\newcommand{\etens}[1]{\mathsfit{#1}}

\def\etA{{\etens{A}}}

\newcommand{\E}{\mathbb{E}}

\newcommand{\R}{\mathbb{R}}

\newcommand{\KL}{D_{\mathrm{KL}}}
\newcommand{\Var}{\mathrm{Var}}

\newcommand{\Cov}{\mathrm{Cov}}

\newcommand{\normltwo}{L^2}
\newcommand{\normlp}{L^p}

\newcommand{\parents}{Pa} 

\usepackage{hyperref}
\usepackage{url}

\usepackage{multirow}
\usepackage{algorithm}
\usepackage{algorithmic}
\usepackage{array}
\usepackage{graphicx}
\usepackage{listings}
\usepackage[np, autolanguage]{numprint}
\usepackage{enumitem}
\usepackage{booktabs}
\usepackage{soul}
\usepackage[skip=0pt]{caption}
\usepackage{subfigure}
\usepackage{amsmath}
\usepackage{amssymb}
\usepackage{pifont}
\usepackage{xspace}
\usepackage{bbding}
\usepackage{wrapfig}
\usepackage{fancyhdr}
\usepackage{makecell}
\usepackage{amsfonts}
\newcommand{\paratitle}[1]{\vspace{1.5ex}\noindent\textbf{#1}}
\newcommand{\ie}{\emph{i.e.,}\xspace}

\newcommand{\eg}{\emph{e.g.,}\xspace}

\newcommand{\ignore}[1]{}

\title{Mix-CPT: A Domain Adaptation Framework via Decoupling Knowledge Learning and Format Alignment}

\author{Jinhao Jiang\textsuperscript{{1}}, Junyi Li\textsuperscript{{1}}, Wayne Xin Zhao\textsuperscript{{1}}\thanks{\llap{}\:Corresponding authors.}, Yang Song\textsuperscript{{3}}\footnotemark[1], Tao Zhang\textsuperscript{{3}}~and Ji-Rong Wen\textsuperscript{{1},{2}} \\
\textsuperscript{1}Gaoling School of Artificial Intelligence, Renmin University of China.\\
\textsuperscript{2}School of Information, Renmin University of China.\\
\textsuperscript{3}BOSS Zhipin, Beijing, China\\
\texttt{jiangjinhao@ruc.edu.cn, lijunyi@ruc.edu.cn,} \\
\texttt{batmanfly@gmail.com, jrwen@ruc.edu.cn}
}

\iclrfinalcopy 
\begin{document}

\maketitle

\begin{abstract}
Adapting general large language models (LLMs) to specialized domains presents great challenges due to varied data distributions. This adaptation typically requires continual pre-training on massive domain-specific corpora to facilitate knowledge memorization, followed by training to apply this knowledge following human instructions and preferences. However, this method may result in inefficient knowledge memorization due to a lack of awareness of knowledge utilization and imposes substantial demands on LLMs to simultaneously learn knowledge utilization and format alignment with limited training samples.
To facilitate the domain adaptation of LLM, we revise this process and propose a new domain adaptation framework including domain knowledge learning and general format alignment, called \emph{Mix-CPT}. Specifically, we first conduct knowledge mixture continual pre-training that concurrently focuses on knowledge memorization and utilization, allowing for mutual reinforcement. To avoid catastrophic forgetting during the continual pre-training process, we further incorporate a logit swap self-distillation constraint. Subsequently, leveraging the knowledge and capabilities acquired during continual pre-training, we efficiently perform instruction tuning and alignment with a few general training samples to achieve format alignment.
Extensive experiments demonstrate that our proposed \emph{Mix-CPT} framework can simultaneously improve the task-solving capabilities of LLMs on the target and general domains compared to the traditional adaptation methods.

\end{abstract}

\section{Introduction}
\label{sec:intro}

Large Language Models (LLMs)~\citep{llm_survey} have revolutionized the field of natural language processing (NLP)~\citep{gpt3,OpenAI-OpenAI-2023-GPT-4}, showing exceptional capabilities such as instruction following~\citep{Ouyang-arxiv-2022-Training,Taori-github-2023-Stanford} and complex reasoning~\citep{cot,self-consistency}. 
However, due to their limited exposure to relevant data, such general LLMs still considerably lag behind in specific domains requiring professional knowledge. This situation has necessitated the effective adaptation of general-purpose LLMs to specific domains (\eg mathematics and code), called \emph{domain adaptation} of  LLMs~\citep{abs-2211-03154}. 

In essence, tailoring general LLMs to specific domains requires adaptation in two main aspects, namely \emph{knowledge learning} (acquiring and leveraging the necessary domain knowledge) and \emph{format alignment} (responding to the user in an expected output form)~\citep{jiang2024instruction, LIMA}. Specially, knowledge learning can be further fulfilled via knowledge memorization and utilization. In practice, domain adaptation of LLMs typically involves three consecutive stages~\citep{CodeLLaMA, Llemma}, \ie pre-training, instruction tuning, and alignment, where the first stage is primarily aimed at knowledge memorization and the other two stages are mainly focused on knowledge utilization and format alignment. However, at the pre-training stage,  knowledge memorization based on raw domain-specific corpus would be somehow inefficient without eliciting the acquired knowledge according to task goals~\citep{jiang2024instruction}. 
Despite that some studies incorporate instruction data for pre-training, they often rely on proprietary models to synthesize high-quality instructions at  scale~\citep{cheng2024instruction,abs-2403-02756}, which may not be that easy without extensive fine-tuning experiences. Another issue is that learning to master knowledge utilization and format alignment in the instruction tuning and alignment stages might lead to suboptimal performance, due to the fact that the two goals can be divergent in model optimization~\citep{Ren-2024-learning}.

\begin{figure}[t]
    \centering
    \includegraphics[width=\textwidth]{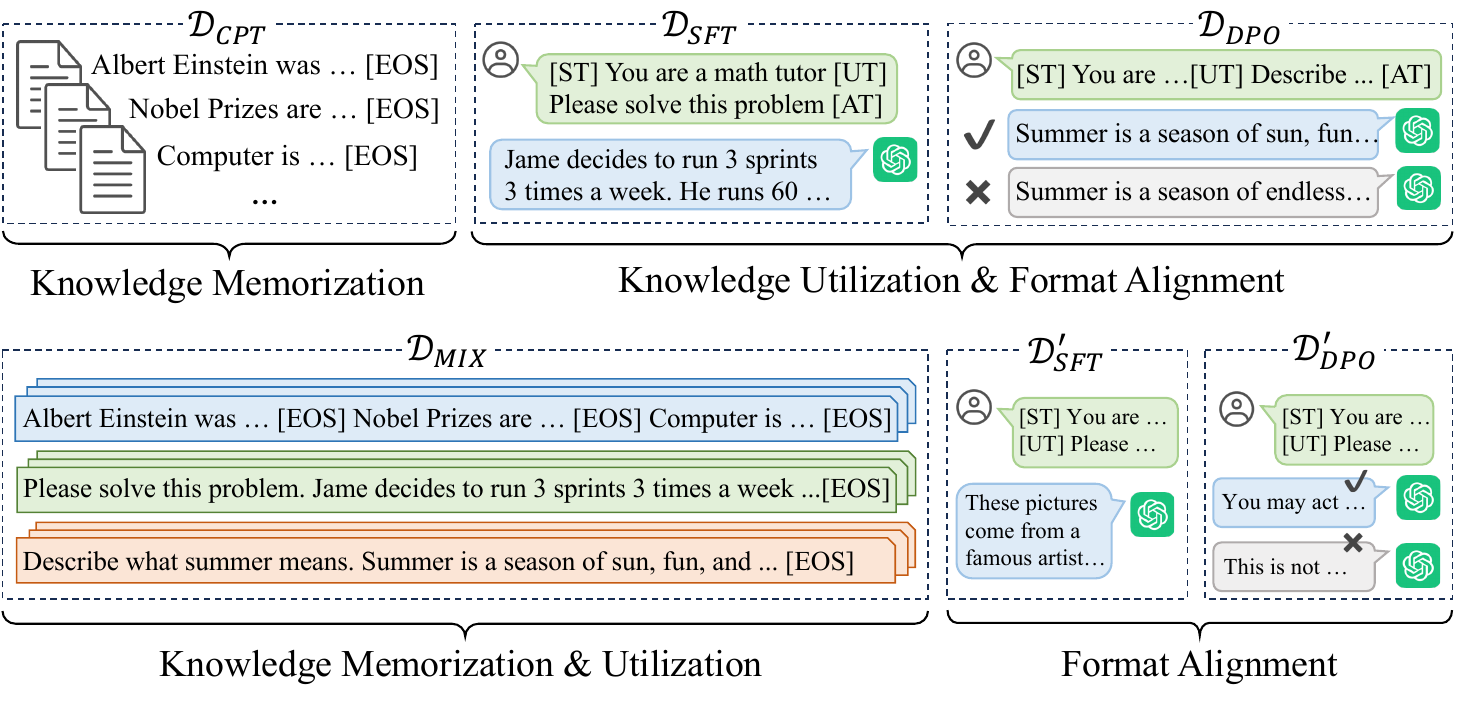}
    \caption{Comparison of traditional domain adaptation approaches~(top) and our proposed rescheduled domain adaptation paradigm~(bottom). ``[EOS]'' is the special token representing the end of the document. ``[ST]'', ``[UT]'' and ``[AT]'' denote the system, user, assistant chat template, repesctively.} 
    \label{fig:data_format}
\end{figure}

Considering the above issues, this paper explores a new domain adaptation approach that only uses raw domain-specific corpus and general instruction or alignment data. 
{Our hypothesis is that the knowledge utilization capacity can be essentially learned from general instruction or alignment data, which has been also evidenced by prior studies~\citep{InstructGPT,DPO}.}
In this way, we can remove the tedious instruction synthesis step from the training pipeline, since it is much easier to obtain general or mixed-domain instruction data from open resources.  
Another important attempt is to enhance knowledge learning by jointly attaining both memorization and utilization of knowledge.
To implement this idea, we schedule all the instruction and alignment data at the pre-training stage (with a suitable format), then only reuse a minor proportion of instruction and alignment data for the instruction-tuning and alignment stages to achieve format alignment. We compare our rescheduled process with the traditional domain adaptation in Figure~\ref{fig:data_format}.

Specially, our approach for domain adaptation of LLMs consists of two main stages, including \emph{domain knowledge learning} and \emph{general format alignment}. In the first stage, we conduct knowledge mixture continual pre-training to integrate both knowledge memorization and utilization. The memorization of new knowledge can be facilitated by taking into account how this knowledge will be utilized. In the second stage, based on the knowledge and capabilities that are already acquired during pre-training, we perform instruction tuning and alignment in an efficient manner to achieve format alignment.
For unified training, we convert raw domain documents, instruction tuning data, and alignment data {into a unified format for conducting knowledge mixture continual pre-training~(\emph{Mix-CPT})}. 
To avoid catastrophic forgetting in continual pre-training, we propose Logit Swap Self-Distillation (LSSD), which exchanges the predicted top-$1$ token logit with the logit of the ground-truth token, serving as the surrogate target. In this way, LSSD maintains most probabilities of the original distribution of LLMs, avoiding dramatic model update and thereby preserving original capabilities. In instruction tuning and alignment, we select a small number of \emph{easy} instructions from the pre-training instruction set based on the perplexity scores of LLMs as criteria. These instructions have already been seen during pre-training, so that the model can mainly focus on pure style or format learning for downstream tasks.

To verify the effectiveness of our proposed Mix-CPT method, we evaluate it on both domain-specific and general tasks, including a total of seven distinctive capabilities based on 17 representative benchmarks. 
For both base LLMs and chat LLMs, our approach can effectively improve their domain-specific and general performance simultaneously compared to traditional methods of first performing continual pre-training, followed by instruction tuning and alignment.
\section{Approach}

\subsection{Overview}

\begin{figure}[t]
    \centering
    \includegraphics[width=\textwidth]{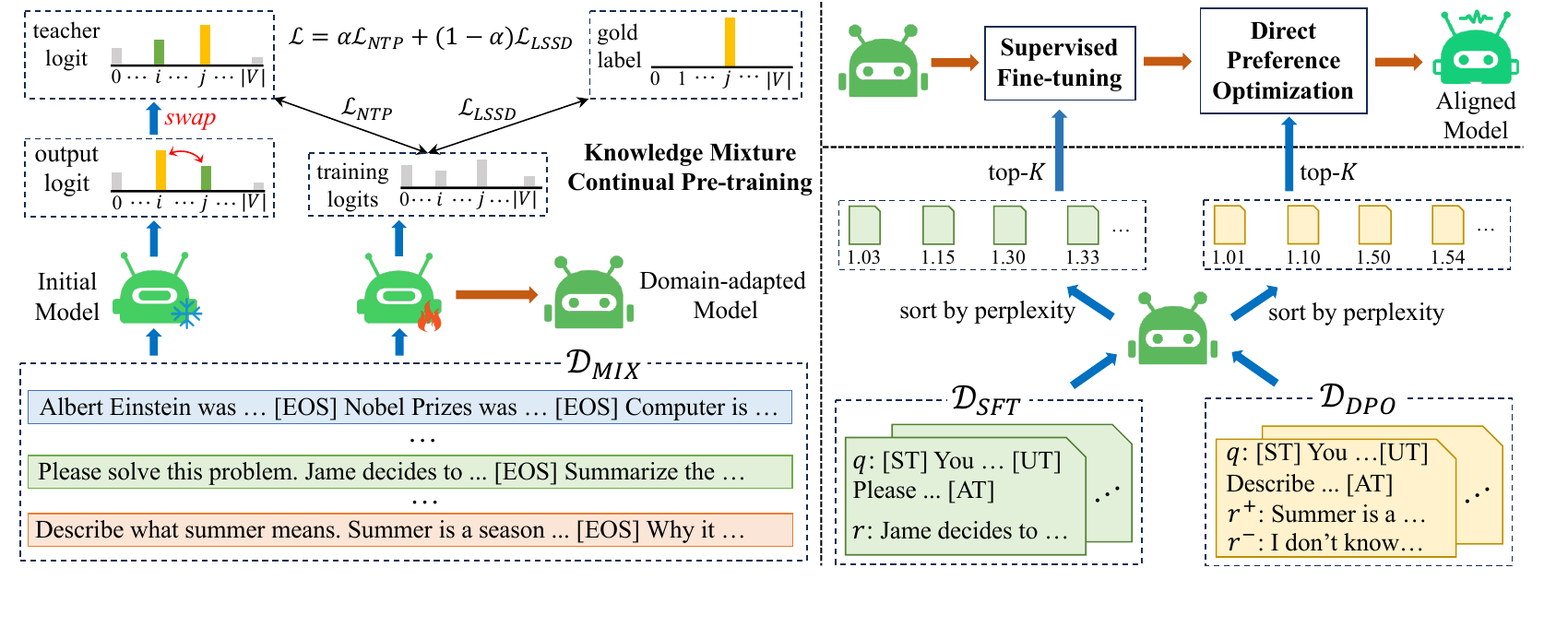}
    \caption{The illustration of our proposed rescheduled domain adaptation paradigm, including first conducting knowledge mixture continual pre-training, then selecting top-$K$ easy training samples with the lowest perplexity for performing supervised fine-tuning and direct preference optimization.} 
    \label{fig:Mix_CPT}
\end{figure}

To adapt general-purpose LLMs to specific domains (\eg Wiki, mathematics, code), our core idea is to decouple knowledge learning and format alignment, and propose an effective two-stage domain adaptation framework for general LLMs, \ie first performing knowledge mixture continual pre-training~(Section~\ref{sec:cpt}) and then performing efficient format alignment~(Section~\ref{sec:alignment}). We show the overall architecture in Figure~\ref{fig:Mix_CPT}.

In the first stage, we conduct continual pre-training on the mixed data of raw domain-related documents, general instruction and alignment data via a unified text format. We aim to utilize general instruction data to better elicit the capacities of knowledge memorization and utilization during continual pre-training. 
To avoid catastrophic forgetting in pre-training, we design a new learning method, \ie Logit Swap Self-Distillation~(LSSD), {that exchanges the top-$1$ token logit with the ground-truth token logit}. In the second stage, based on the domain knowledge augmented LLMs, we conduct efficient format alignment with a small number of easy instructions or preference samples that have been seen during pre-training. In this way, LLMs can focus on learning the simple style and formet for interacting with human, without much consideration of how to utilize the attained knowledge. Next, we will describe each part in detail.

\subsection{Knowledge Mixture Continual Pre-training}
\label{sec:cpt}
Different from prior work that performs continual pre-training solely based on domain-specific corpus~\citep{que2024d,KeSL00022}, we propose to mix domain-specific documents, general instructions and alignment data as pre-training data. 
The QA-based instruction data is useful to reflect how the knowledge will be accessed and utilized through questions. 
In this way, LLMs can improve their capabilities to learn new knowledge from those domain-related documents. 
Incorporating general instructions can facilitate LLMs to transfer the general knowledge utilization capability to domain-specific knowledge without relying on highly domain-related instructions in previous work~\citep{jiang2024instruction,cheng2024instruction}. 
Specifically, we first transform the raw domain documents, general instructions and alignment data into a unified knowledge format. Then, we perform continual pre-training on this mixture collection with the objective of next token prediction. To avoid catastrophic forgetting, we further introduce a logit swap self-distillation approach during the continual pre-training process. Next, we introduce these techniques in detail. 

\subsubsection{Unified Knowledge Format}
Typically, adapting a base LLM to a specific domain involves three distinct and relatively independent stages, each based on corresponding data in different formats. 
Specifically, the base model firstly performs continual pre-training (CPT) on domain-specific corpus for learning new knowledge, then conducts supervised fine-tuning (SFT) based on instructions for enhancing the instruction following ability, and finally utilizes the human preference data for human alignment.
In this work, we adopt the direct preference optimization (DPO) to as the alignment algorithm. 
Formally, we denote the domain-specific corpus as $\mathcal{D}_\text{CPT} = \{{d_i}\}_{i=1}^{n_c}$, where ${d_i}$ represents a raw domain document consisting of a sequence of tokens. For the instructions used in SFT, we denote as $\mathcal{D}_\text{SFT} = \{\langle q_i, r_i \rangle \}_{i=1}^{n_s}$, where ${q_i}$ and ${r_i}$ represent the user query and the expected response, repectively. For alignment data used in DPO, we denote by $\mathcal{D}_\text{DPO} = \{\langle q_i, r^+_i, r^-_i \rangle \}_{i=1}^{n_d}$, where ${q_i}$, ${r}^+_i$, and ${r}^-_i$ represent the user query, positive response, and negative response, respectively.

In this work, we propose to mix $\mathcal{D}_\text{CPT}$, $\mathcal{D}_\text{SFT}$ and $\mathcal{D}_\text{DPO}$ with a unified text format, building upon which we further perform knowledge mixture continual pre-training on a general LLM. Unlike previous work relying on synthesizing high-quality domain instructions~\citep{jiang2024instruction,cheng2024instruction}, we empirically find that knowledge utilization 
is indeed a general capability that can be learned from general instructions, and we can further transfer such capacity to enhance the learning of domain knowledge.
Specially, we remove any templates and markers (\eg \texttt{[User]}) from the instructions and alignment data to construct the mixture data in a unified format, denoted by $\mathcal{D}_\text{MIX} = \{ x_\text{cpt}, x_\text{sft}, x_\text{dpo} \}$, where $x_\text{cpt} = d_i$ is the original domain document, $x_\text{sft} = [{q_i};{r_i}]$ denotes the concatenation of user query and expected response in instructions, and $x_\text{dpo} = [{q_i};{r}^+_i]$ is the concatenation of user query and positive response in the alignment data. We show some examples in Figure~\ref{fig:data_format}.

{Following existing pre-training methods~\citep{LLaMA}}, during the continual pre-training process, we concatenate each kind of data sample (\ie $x_\text{cpt}$, $x_\text{sft}$, or $x_\text{dpo}$) and truncate the sequence when reaching the maximum input length of the LLM. Besides, we add an extra special symbol (\ie \texttt{[SEP]}) at the end of each sample to separate them. We repeat this process until concatenating all samples from each kind of data (\ie $\mathcal{D}_\text{CPT}$, $\mathcal{D}_\text{SFT}$ and $\mathcal{D}_\text{DPO}$) to obtain our final knowledge mixture pre-training data $\mathcal{D}_\text{MIX}$. 

Note that though we use general instruction data to derive the mixture data here, our approach can be generally extended to incorporating domain-specific instruction data~\citep{cheng2024instruction}, which often relies on specific data synthesis techniques. 

\subsubsection{Logit Swap Self-Distillation}
After obtaining the mixture data $\mathcal{D}_\text{MIX}=\{x_\text{cpt}, x_\text{sft}, x_\text{dpo}\}$, we then perform continual pre-training on the base LLM. For simplicity, we remove the subscript of each training sample in the mixture data, denoted as $x$. We adopt the pre-training task of next token prediction (NTP), which aims to predict the next token conditioned on all previous tokens. Specifically, given an input $x=\{w_1, w_2, ..., w_n\}$, we feed it into the decoder-only LLM and use the standard language modeling objective to minimize the cross-entropy loss as follows:
\begin{align}
    \mathcal{L}_\text{NTP} = - \sum_{j=1}^n \log \text{Pr}(w_j|w_{<j};\Theta),
\end{align}
where $w_j$ denotes the $j$-th token in the input, $w_{<j}$ is the previous tokens, and $\Theta$ denotes the model parameters. During continual pre-training, the task of next-token prediction enables the base LLM to learn domain knowledge, and the incorporation of general instruction and alignment data further transfers the general knowledge utilization capability to specific domains. 

However, the traditional language modeling objective is prone to suffer from the issue of catastrophic forgetting for previously learned knowledge of LLMs. Therefore, we propose an auxiliary training objective, \ie \emph{Logit Swap Self-Distillation (LSSD)}, which serves as an extra constraint for standard language modeling objective. Specifically, we first utilize the original base LLM before continual pre-training (paramerized by $\Theta_\text{ori}$) to infer the output logits following the standard language modeling objective yet without computing the loss:
\begin{align}
    \bm{h}_j &= \text{LLM}(w_{<j};\Theta_\text{ori}),\\
    \bm{l}_{j} &= \bm{h}_j\bm{W}_e^T,
\end{align}
where $\bm{W}_e$ is the token embedding matrix, $\bm{h}_j$ is the hidden state of the last transformer block, and $l_j$ denotes the output logit at the $j$-th position.
Then, we exchange the logit value of the top-$1$ predicted token (\ie $\widetilde{w}_j$) and the ground-truth token (\ie $w_j$) if they are not equal as follows:
\begin{align}
    \bm{\tilde{l}}_j = \text{Exchange}(\bm{l}_j, I_{\widetilde{w}_j}, I_{w_j}), \quad \text{if}~~ I_{\widetilde{w}_j} \neq I_{w_j},
\end{align}
where $I_{\widetilde{w}_j}$ and $I_{w_j}$ denote the indices of the top-$1$ predicted token $\widetilde{w}_j$ and the ground-truth token $w_j$ in the vocabulary, respectively, the function Exchange($\cdot$) will exchange their logit values in $\bm{l}_j$, and the output $\bm{\tilde{l}}_j$ will be regarded as the teacher logit in LSSD. In essence, LSSD only calibrates the prediction of ground-truth token for adapting to the current domain knowledge while maintaining the most previously learned knowledge of the base LLM (\ie represented by the unchanged logit values in $\bm{\tilde{l}}_j$). 
Then, we can compute the teacher model's probability distribution for the $j$-th token with softmax function:
\begin{align}
\text{Pr}(w_j|w_{<j};\Theta_\text{ori}) = \text{softmax}(\bm{\tilde{l}}_j),
\end{align}
Finally, we compute the self-knowledge distillation objective and minimize the reverse Kullback-Leibler divergence loss~\citep{MiniLM} between the current model's probability distribution and the teacher model's probbaility distribution as follows:
\begin{align}
    \mathcal{L}_\text{LSSD} = - \sum_{j=1}^n\sum_{w_j \in \mathcal{V}}\text{Pr}(w_j|w_{<j};\Theta)\log(\frac{\text{Pr}(w_j|w_{<j};\Theta)}{\text{Pr}(w_j|w_{<j};\Theta_\text{ori})}),
\end{align}
where $\mathcal{V}$ is the vocabulary and $\Theta$ denotes the parameters of the current LLM. In the knowledge mixture continual pre-training stage, the final total loss is the combination of next token prediction loss and self-distillation loss as follows:
\begin{align}
    \mathcal{L}_\text{CPT} = \alpha \cdot \mathcal{L}_\text{NTP} + (1-\alpha) \cdot \mathcal{L}_\text{LSSD},
\end{align}
where $\alpha$ is a coefficient to control the proportion of two parts.

\subsection{Efficient Format Alignment}
\label{sec:alignment}

In the domain knowledge learning stage, the LLM has simultaneously learned to both memorize domain knowledge and understand how to utilize the knowledge through our proposed knowledge mixture continual pre-training. After that, during the format alignment stage, the LLM can more efficiently fine-tuned to master the task format with only a small number of alignment samples. Next, we first introduce the selection of training samples and then perform general format alignment. 

Since we would like to decouple knowledge learning and format alignment, we mainly focus on training samples from $\mathcal{D}_\text{SFT}$ and $\mathcal{D}_\text{DPO}$ that are are both \emph{easy} and have been \emph{encountered} during continual pre-training, which avoids introducing new knowledge during supervised fine-tuning. 
These easy samples are selected based on the perplexity scores of the LLMs \emph{w.r.t} the ground-truth output. Specifically, given a sample in $\mathcal{D}_\text{SFT}$ and $\mathcal{D}_\text{DPO}$, we first equip the input instruction and output response with the corresponding chat template, \ie the format for interaction with humans. Then, we feed the formatted sequence into the LLM and compute the perplexity score for the output response. Finally, we select top-$K$ samples with the lowest perplexity scores to conduct the supervised fine-tuning and direct preference optimization. Note that for samples in $\mathcal{D}_\text{DPO}$, we only utilize the positive response for computing its perplexity score.

After selecting the training samples, we next utilize them to conduct efficient format alignment. Firstly, we utilize the selected easy instruction samples from $\mathcal{D}_\text{SFT}$ to perform supervised fine-tuning based on the LLM after continual pre-training following the standard way~\citep{InstructGPT}, which is to minimize the cross-entropy loss:
\begin{align}
    \mathcal{L}_\text{SFT} = - \sum_{j=1}^n \log \text{Pr}(r_j|q, r_{<j};\Theta),
\end{align}
where $r_j$ and $r_{<j}$ denote the $j$-th token and its previous tokens in the response.
Secondly, we further utilize the selected easy preference samples from $\mathcal{D}_\text{DPO}$ to conduct direct preference optimization following its original method~\citep{DPO} as follows:
\begin{align}
    \mathcal{L}_\text{DPO} = - \log \sigma \bigg(\beta \log \frac{\pi(r^+|q;\Theta)}{\pi(r^+|q;\Theta_\text{ref})} - \beta \log \frac{\pi(r^-|q;\Theta)}{\pi(r^-|q;\Theta_\text{ref})}\bigg),
\end{align}
where $\sigma$ denotes the sigmoid function, $\Theta$ and $\Theta_\text{ref}$ denote the parameters of the updated LLM and reference LLM during the direct preference optimization process, and $\pi$ denotes the product of the probabilities of all output tokens, conditioned on the given input.

\section{Experiments}

\subsection{Experimental Setup}

\paratitle{Domain-specific Corpus.}
In our experiments, we mainly focus on three popular domains for adapting general-purpose LLMs, \ie encyclopedia, mathematics, and code. For the encyclopedia domain, we select Wikipedia as the primary corpus, which is collaboratively developed by volunteers globally and can be freely accessed online. To enable LLMs to learn knowledge from new documents, we utilize the official 2024/03/01 Wikipedia dump\footnote{https://dumps.wikimedia.org/enwiki/20240301/} and conduct necessary data cleaning and filtering processes such as deduplication, resulting in approximately 4B tokens in raw Wikipedia documents. For the domain of mathematics, we opt for AutoMathText~\citep{AutoMathText}, a carefully curated corpus derived from various sources including websites, arXiv, and GitHub. Each sample in this corpus has been labeled with a quality score from 0.0 (``the poorest'') to 1.0 (``the best''), reflecting its relevance, quality, and educational value in the context of mathematical intelligence. Following previous work~\citep{JiuZhang3.0}, we specifically select those samples with scores higher than 0.7, containing about 0.7B tokens. For the field of code, we select the StarCoder~\citep{StarCoder} corpus, which is widely recognized and employed in several studies~\citep{WizardCoder}. It contains 86 programming languages, and we select the Python subset with approximately 1B tokens.

\paratitle{General Instruction Datasets.}
For general instruction datasets, we choose TULU-V2-mix~\citep{Tulu-V2} and UltraFeedback~\citep{UltraFeedback} for instruction tuning and alignment, respectively. Specifically, each sample in TULU-V2-mix is either manually curated for quality or generated from GPT models for encouraging complexity and diversity. We utilize the entire dataset of TULU-V2-mix (about 326K samples) mixed with domain-specific corpus for knowledge mixture continual pre-training (Section~\ref{sec:cpt}), and then randomly select 10K samples with the lowest perplexity score for subsequent instruction tuning. In addition, UltraFeedback is a widely-used diverse human preference alignment dataset, containing approximately 64K preference pairs. Similarly, we employ the whole dataset of UltraFeedback for knowledge mixture continual pre-training and then downsample 5K pairs with the lowest perplexity score for alignment. It is noted that our TULU-V2-mix and UltraFeedback datasets are open-source and widely used in previous work~\citep{SimPO,SIT}, ensuring a high level of transparency and facilitating fair experimental comparisons.

\paratitle{Baselines.} In the experiments, we employ the Meta-Llama-3-8B~(\emph{LLaMA3-8B})~\footnote{https://llama.meta.com/llama3/} as the base model, which is extensively utilized in the existing work~\citep{cheng2024instruction}. For comparative analysis, we consider the following three types of baseline methods:

~$\bullet$ \emph{Closed-Source Chat LLMs} consist of the official Chat LLMs that have undergone both instruction tuning and preference alignment using closed-source data. Here, we select the Meta-Llama-3-8B-Instruct~(\emph{LLaMA3-8B-Chat}).

~$\bullet$ \emph{Open-Source Chat LLMs} are developed by us following the processes of instruction tuning and alignment. Based on our selected base LLM, we conduct supervised fine-tuning (SFT) using TULU-V2-mix, followed by direct preference optimization (DPO) with UltraFeedback.

~$\bullet$ \emph{Continual Pre-training Augmented LLMs} include domain knowledge-enhanced Chat LLMs which initially undergo continual pre-training~(CPT) with domain-specific corpus, followed by the same implementation of supervised fine-tuning (SFT) and direct preference optimization (DPO) using TULU-V2-mix and UltraFeedback as open-source chat LLMs.

\paratitle{Evaluation Benchmarks and Metrics.} For a comprehensive evaluation, we evaluate seven distinctive capabilities of LLMs based on a total of 17 representative NLP datasets:

$\bullet$ \emph{Factual Question Answering} assesses the factual knowledge of LLMs in the Wikipedia domain. We employ NaturalQuestion~(NQ)~\citep{NQ}, TrivialQA~(TQ)~\citep{TQ}, and our curated WikiQA~(WQ) datasets and use the \emph{Exact Match~(EM)} metric to determine if the prediction is the same as the gold answer.

$\bullet$ \emph{Math Reasoning} tests the LLMs' ability to solve mathematical problems. We use the GSM8K~\citep{GSM8K} and MATH~\citep{MATH} datasets, and evaluate predictions using the \emph{Accuracy} metric.

$\bullet$ \emph{Code Reasoning} tests the LLMs' ability to solve programming problems. We use MBPP~\citep{MBPP} and HumanEval~\citep{HumanEval} datasets, with the \emph{Pass@K} metric assessing the likelihood that at least one of the top-$K$ generated code samples for a problem passes the unit tests.

$\bullet$ \emph{Reading Comprehension} measures the LLMs' ability to comprehend a passage and answer related questions. We use the RACE-Hard~\citep{RACE} and OpenBookQA~\citep{OpenBookQA} datasets, and employ the \emph{Exact Match~(EM)} metric.

$\bullet$ \emph{Commonsense Reasoning} evaluates the ability to answer questions using commonsense knowledge. We use the HellaSwag~\citep{HellaSwag}, CSQA~\citep{CSQA}, and PIQA~\citep{PIQA} datasets, and employ the \emph{Accuracy} metric.

$\bullet$ \emph{Examination} includes comprehensive and challenging benchmarks designed to assess problem-solving ability across various domains. We use MMLU~\citep{MMLU}, BBH~\citep{BBH}, and ARC-Challenge~\citep{ARC} for English examinations and C-EVAL~\citep{C-Eval} for Chinese. Both benchmarks are evaluated using the \emph{Accuracy} metric.

$\bullet$ \emph{Instruction Following} assesses the LLMs' ability to engage in coherent, informative, and engaging conversations. We use the MT-Bench~\citep{MT-Bench} datasets. For evaluation, we utilize the GPT-4-0613~\footnote{https://platform.openai.com/docs/models} as the judging model, assigning a score ranging from 1 to 10 to the answer. We multiply this score by ten, resulting in a final score of 100.

Specifically, we evaluate the above datasets based on the OpenCompass framework~\citep{OpenCompass}, which is a one-stop platform for large model evaluation, aiming to provide a fair, open, and reproducible benchmark for large model evaluation.

\subsection{Main Results}
Table~\ref{tab:base_res} and Table~\ref{tab:chat_res} display the evaluation results of our proposed Mix-CPT framework and other baselines using specific evaluation benchmarks. In the next, we give a detailed analysis. 
 
\subsubsection{Results of Base LLMs}
We initially assess the effectiveness of our proposed Mix-CPT framework, especially the logit swap self-distillation constraint, in mitigating catastrophic forgetting and facilitating the knowledge learning of base LLMs during the continual pre-training stage.
To better observe the impact of knowledge mixture continual pre-training on the performance in target domains and general capabilities, we select domain-specific tasks and comprehensive examination tasks.
Specifically, we evaluate the domain-specific tasks (\eg factual question answering~(Wiki), math reasoning~(Math), and code reasoning~(Code)) under few-shot settings and the Chinese and English examination tasks (\ie multi-choice question answering) with perplexity-based zero-shot setting. We show the evaluation results in Table~\ref{tab:base_res}.

First, we can see that traditional continual pre-training (\ie + CPT) on the raw domain data does not necessarily enhance the performance of base LLMs in the target domain, and may instead impair their performance therein. Additionally, this method inevitably leads to a certain degree of catastrophic forgetting, thereby damaging the overall performance of the base LLM.
For example, when compared to the LLaMA3-8B-Base model, performing continual pre-training in the Wiki or Math domains leads to observed improvements in the corresponding target domains~(\ie 50.48 $\rightarrow$ 50.79 in the Wiki domain and 33.17 $\rightarrow$ 37.72 in the Math domain). However, this approach is inefficient in the Code domain, where it results in diminished performance~(\ie 58.75 $\rightarrow$ 55.64 in Code domain). 
The phenomenon is also discovered in existing work~\citep{Rho}, which suggests that the data used for continual pre-training should be of superior quality relative to the data utilized during the initial pre-training phase; otherwise, it may detrimentally affect performance. Indeed, the publicly available domain data we utilized~(\eg AutoMathText and StarCoder) has likely already been used to train the base model. Consequently, further employing it for continual pre-training may impair the model's performance.

Second, when mixing the domain raw data with the additional instructions and alignment data (\ie Mix-CPT w/o KD), the domain-specific capability can be further improved, which indicates that the mixed instruction data can benefit the learning of the domain knowledge during continual pre-training. At the same time, it can mitigate the effect of other general capabilities and reduce the degradation of the overall average performance.
For example, compared to the traditional continual pre-training method~(CPT), mixing domain corpus with additional instruction data~(Mix-CPT w/o KD) can almost consistently improve the domain capability and overall average performance~(\ie 55.64 $\rightarrow$ 57.98 and 43.94 $\rightarrow$ 45.16 in the target Code domain and overall average performance).

Finally, through applying the logit swap self-distillation strategy to the knowledge mixture continual pre-training process~(\ie Mix-CPT), we can further reduce the impact on the pre-learned knowledge for the base LLM while maintaining the domain capability improvement obtained from simple knowledge mixture~(\ie Mix-CPT w/o KD), thereby mitigating the degradation of the general capabilities of LLMs. Therefore, these results demonstrate that the Mix-CPT framework with the logit swap self-distillation constraint can indeed promote knowledge learning and alleviate the issue of catastrophic forgetting to some extent.

\begin{table*}[t]
\setlength\tabcolsep{5pt}
\centering
\small
\caption{Evaluation results on three specialized domains and two comprehensive examination domains. The \underline{underline} and \textbf{bold} fonts denote the best results in the target domain and the average performance in each domain adaptation group, respectively.}
\label{tab:base_res}
\begin{tabular}{l l c c c c c c}
\toprule
\multicolumn{2}{c}{\textbf{Model}} & \textbf{Wiki} & \textbf{Math} & \textbf{Code} & \textbf{\makecell{English\\ Examination}} & \textbf{\makecell{Chinese\\ Examination}} & \textbf{Average} \\
\midrule
\multicolumn{2}{l}{LLaMA3-8B-Base} & 50.48 & 33.17 & 58.75 & 50.93 & 47.95 & 45.72    \\
\cmidrule{3-8}
\multirow{3}{*}{\textit{Wiki}} & + CPT & 50.79 & 34.02 & 15.56 & 52.79 & 46.27 & 42.00    \\
 & + Mix-CPT (\emph{w/o KD}) & \underline{52.47} & 39.14 & 57.20 & 52.25 & 47.64 & 47.59    \\
 & + Mix-CPT & 52.25 & 38.13 & 56.42 & 52.58 & 49.83 & \textbf{47.73}  \\
\cmidrule{3-8}
\multirow{3}{*}{\textit{Math}} & + CPT & 50.60 & 37.72 & 54.47 & 44.13 & 46.78 & 44.06   \\
 & + Mix-CPT (\emph{w/o KD}) & 50.60 & \underline{37.73} &54.47 &43.47 &48.77 & 44.78  \\
 & + Mix-CPT & 50.70 & 37.67 & 57.98 & 51.45 & 49.93 & \textbf{47.51}   \\
\cmidrule{3-8}
\multirow{3}{*}{\textit{Code}} & + CPT & 46.14 & 33.71 & 55.64 & 51.38 & 47.58 & 43.94   \\
 & + Mix-CPT (\emph{w/o KD}) & 50.09 & 37.85 & 57.98 & 46.82 & 47.91 & 45.16   \\
 & + Mix-CPT & 50.73 & 38.01 & \underline{59.14} & 51.49 & 47.87 & \textbf{46.90}   \\
\bottomrule
\end{tabular}
\end{table*}

\subsubsection{Results of Chat LLM}
Subsequently, we assess the performance of final chat LLMs after performing format alignment with selectively chosen training samples for instruction tuning and alignment. We want to examine the instruction following capabilities of these LLMs using comprehensive benchmarks in a zero-shot scenario. Accordingly, we assess both domain-specific tasks~(\ie factual question answering~(Wiki), math reasoning~(Math), and code reasoning~(Code)) and more general tasks~(\ie Reading Comprehension~(RC), Commonsense Reasoning~(CR), Examination~(EX), and Instruction Following~(IF)). We show the final results in Table~\ref{tab:chat_res}.

First, the traditional domain adaptation method~(\ie CSD consisting of CPT, SFT, and DPO), faces challenges in simultaneously enhancing domains-specific capabilities while preserving general capabilities, in contrast to 
open-source chat models that do not utilize domain-specific raw data.
For example, compared to open-source LLaMA3-8B chat model in the Wiki domain, traditional methods of conducting an extra continual pre-training based on raw Wikipedia documents not only decreases the factual question answering performance (\ie 26.10 $\rightarrow$ 22.20 in Wiki domain), but also hurts the average performance (\ie 52.57 $\rightarrow$ 50.96 in overall average performance).
In the Math and Code domains, despite obtaining improvements in the target domains~(\ie 30.83 $\rightarrow$ 37.35 in Math domain and 42.53 $\rightarrow$ 48.55 in Code domain), the overall capability of LLMs exhibited a decline~(\ie 52.57 $\rightarrow$ 52.10 after adapting to the Math domain).
This phenomenon can also be observed in the other two LLMs, which indicates that conventional domain adaptation methods of performing continual pre-training on raw domain data may cause catastrophic forgetting and merely focus on knowledge memorization without considering how to utilize the learned knowledge, which might suffer from the memorization trap.

Second, our proposed Mix-CPT method can simultaneously improve the performance of the target domains and the general capability. The main reasons are two fold. On one hand, with the constraint of logit swap self-knowledge distillation during continual pre-training, the LLM can effectively memorize the raw domain data while maintaining its originally learned knowledge in the previous pre-training stage. On the other hand, by mixing the raw domain data with the general instruction and alignment data (removing any templates), the model can learn the general knowledge utilization capability and transfer this capability to utilize the raw domain data. In this way, the model can perform efficient format alignment with only a few formatted samples to better utilize both target domain knowledge and other general knowledge.

Finally, with the same instruction data and alignment data, our method can successfully improve the performance on the target domain while maintaining the general capability compared to the traditional continual pre-training augmented method based on the raw domain data. Compared to the official chat LLM with large-scale closed-source instruction tuning and alignment tuning, our proposed method can achieve comparable even better performance on the target domain (\eg 40.28 vs 44.09 in the Wiki domain), which also indicates the effectiveness of our method. 

\begin{table*}[t]
\setlength\tabcolsep{4.8pt}
\centering
\small
\caption{Evaluation results on three domain capabilities (\ie Wiki, Math, and Code) and four general capabilities (\ie Reading Comprehension, Complex Reasoning, EXamination, and Instruction Following) with the average performance. The \underline{underline} and \textbf{bold} fonts denote the best results in the target domain and the average performance in each domain adaptation group, respectively.}
\label{tab:chat_res}
\begin{tabular}{lllccc ccccc}
\toprule
\multicolumn{3}{c}{\multirow{2.5}{*}{\textbf{Model}}} & \multicolumn{3}{c}{\textbf{Specialized Domain}} & \multicolumn{4}{c}{\textbf{General Domain}} & \multirow{2.5}{*}{\textbf{Average}}  \\
\cmidrule(r){4-6}\cmidrule(r){7-10}
 & & &  Wiki & Math & Code & RC & CR & EX & IF & \\
\midrule
\multirow{9.5}{*}{\makecell{LLaMA3\\8B-Base}} & \multicolumn{2}{l}{Closed-Source Chat} & 40.28 & 50.52 & 61.54 & 77.23 & 76.59 & 63.60 & 81.04 & 62.62 \\
& \multicolumn{2}{l}{Open-Source Chat} & 26.10 & 30.83 & 42.53 & 73.90 & 75.57 & 56.44 & 68.50 & 52.57 \\
\cmidrule{2-11}
& \multirow{2}{*}{\textit{Wiki}} & + CSD & 22.20 & 30.39 & 40.73 & 74.21 & 73.66 & 56.22 & 63.15 & 50.96 \\
& & + Mix-CPT~(ours) & \underline{44.09} & 31.18 & 48.53 & 70.07 & 68.85 & 57.47 & 70.72 & \textbf{55.23}  \\
\cmidrule{4-11}
& \multirow{2}{*}{\textit{Math}} & + CSD & 12.65 & \underline{37.35} & 44.19 & 74.34 & 77.80 & 58.38 & 69.02 & 52.10  \\
&  & + Mix-CPT~(ours) & 44.35 & 35.99 & 52.55 & 72.88 & 72.91 & 61.64 & 71.31 & \textbf{58.38}  \\
\cmidrule{4-11}
& \multirow{2}{*}{\textit{Code}} & + CSD & 17.27 & 38.13 & 48.55 & 74.75 & 77.82 & 59.21 & 70.88 & 53.87  \\
&  & + Mix-CPT~(ours) & 42.17 & 30.70 & \underline{53.91} & 72.60 & 71.57 & 60.44 & 71.52 & \textbf{56.99}   \\
\bottomrule
\end{tabular}
\end{table*}
\subsection{Detailed Analysis}

\begin{figure}[t]
	\centering
        \begin{minipage}[c]{0.62\textwidth}
		\centering
		\includegraphics[width=\textwidth]{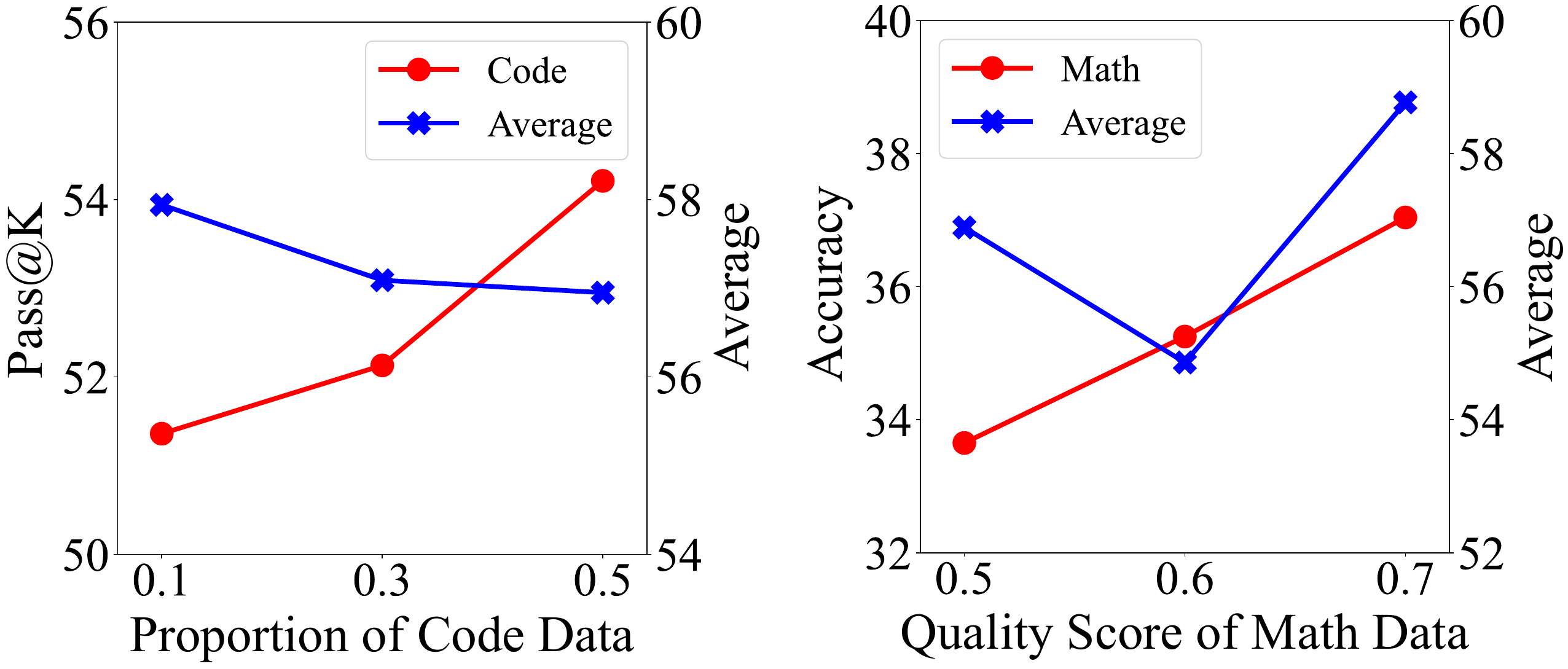}
		\caption{\textbf{(Left)} Pass@K on code and Average results \emph{w.r.t.} Proportion of code data. \textbf{(Right)} Accuracy on math and Average results \emph{w.r.t.} Quality score of math data.}
		\label{fig:code_math_amount}
	\end{minipage}
        \hfill
	\begin{minipage}[c]{0.37\textwidth}
		\centering
		\includegraphics[width=\textwidth]{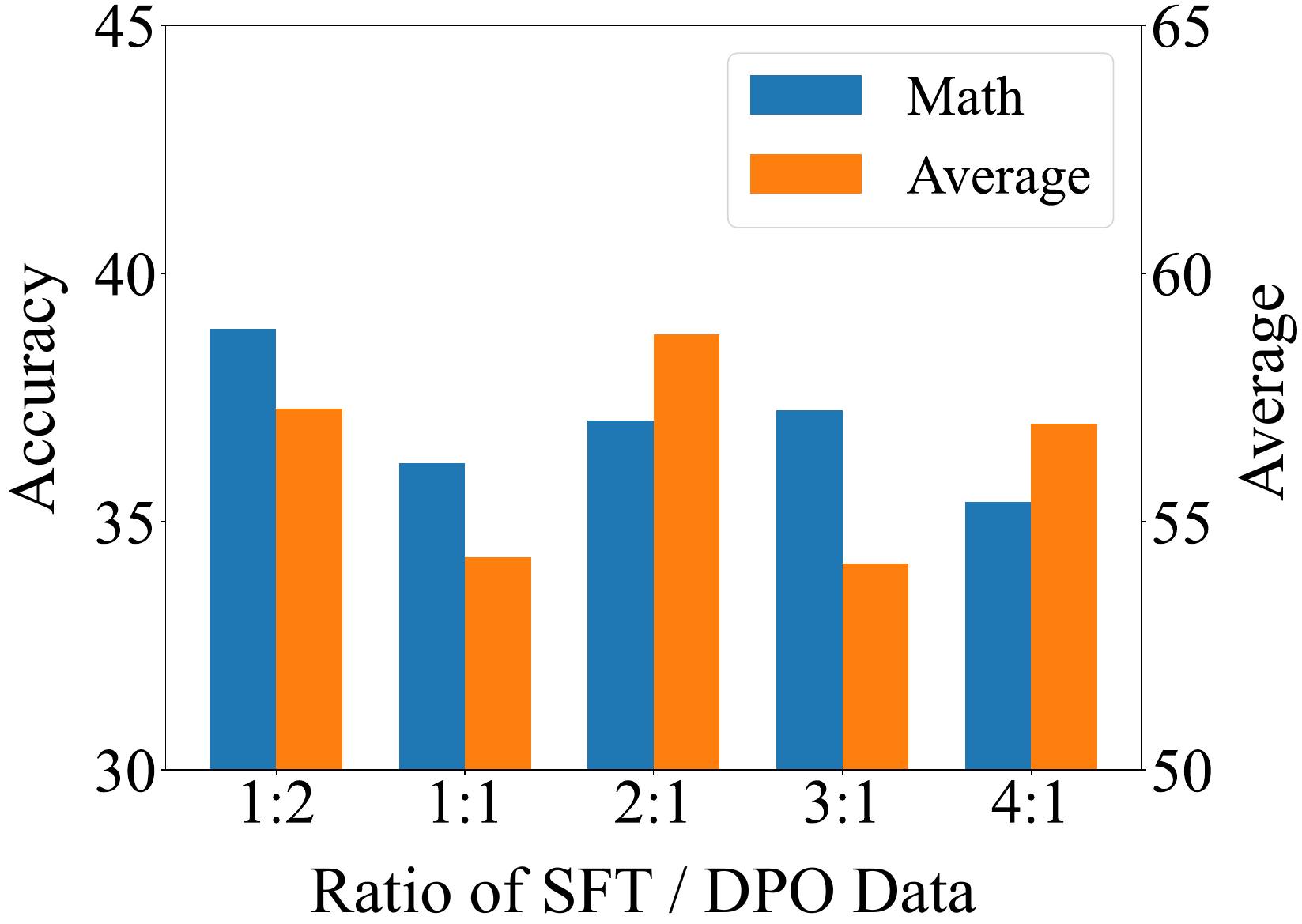}
		\caption{Accuracy on math and Average results \emph{w.r.t.} Ratio of SFT and DPO data.}
		\label{fig:sft_dpo_ratio}
	\end{minipage} 
\end{figure}

\begin{figure}[t]
	\centering
	\begin{minipage}[c]{0.37\textwidth}
		\centering
		\includegraphics[width=\textwidth]{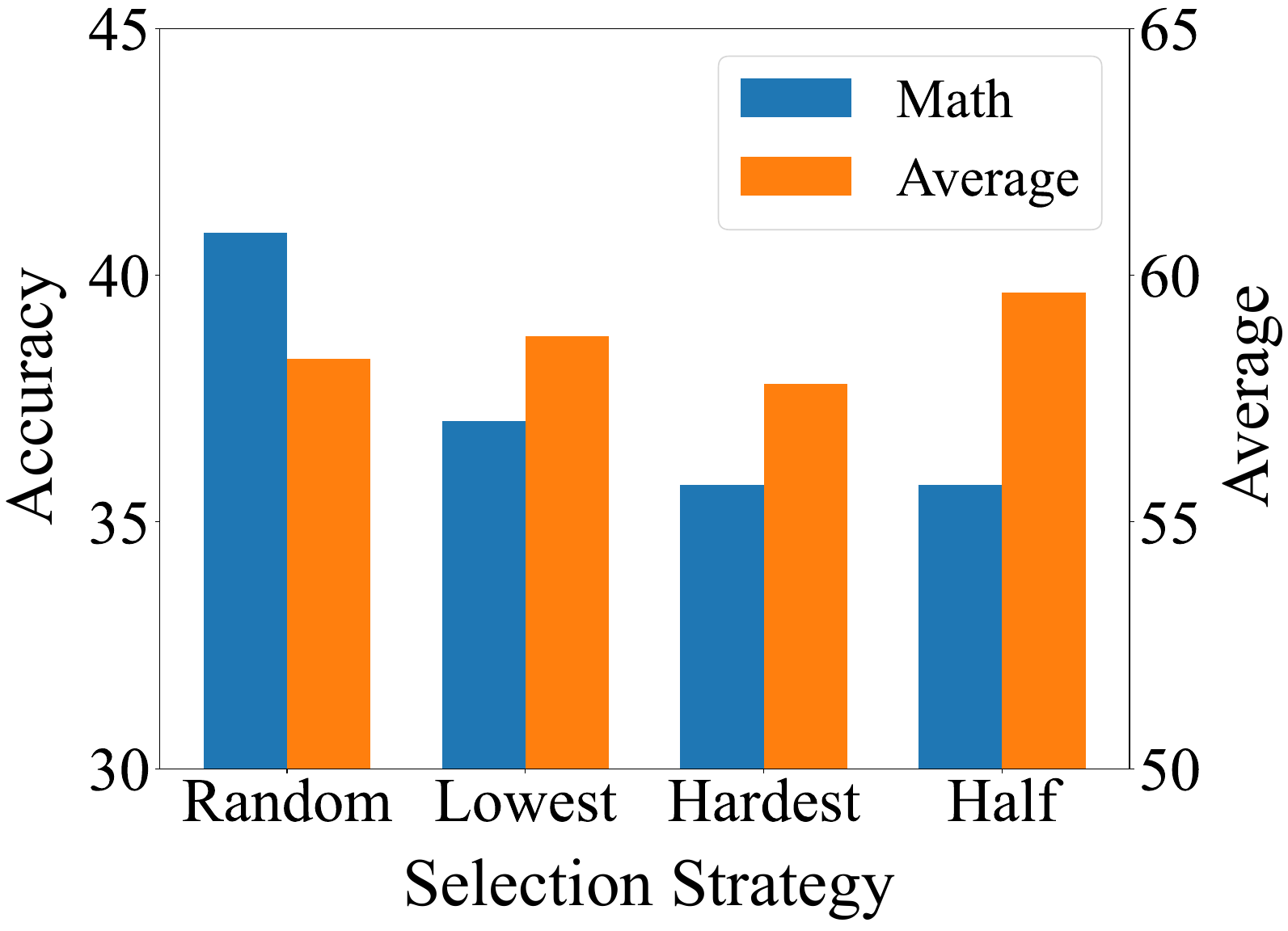}
		\caption{Accuracy on math and Average results \emph{w.r.t.} selection strategy.}
		\label{fig:selection_strategy}
	\end{minipage} 
        \hfill
	\begin{minipage}[c]{0.62\textwidth}
		\centering
		\includegraphics[width=\textwidth]{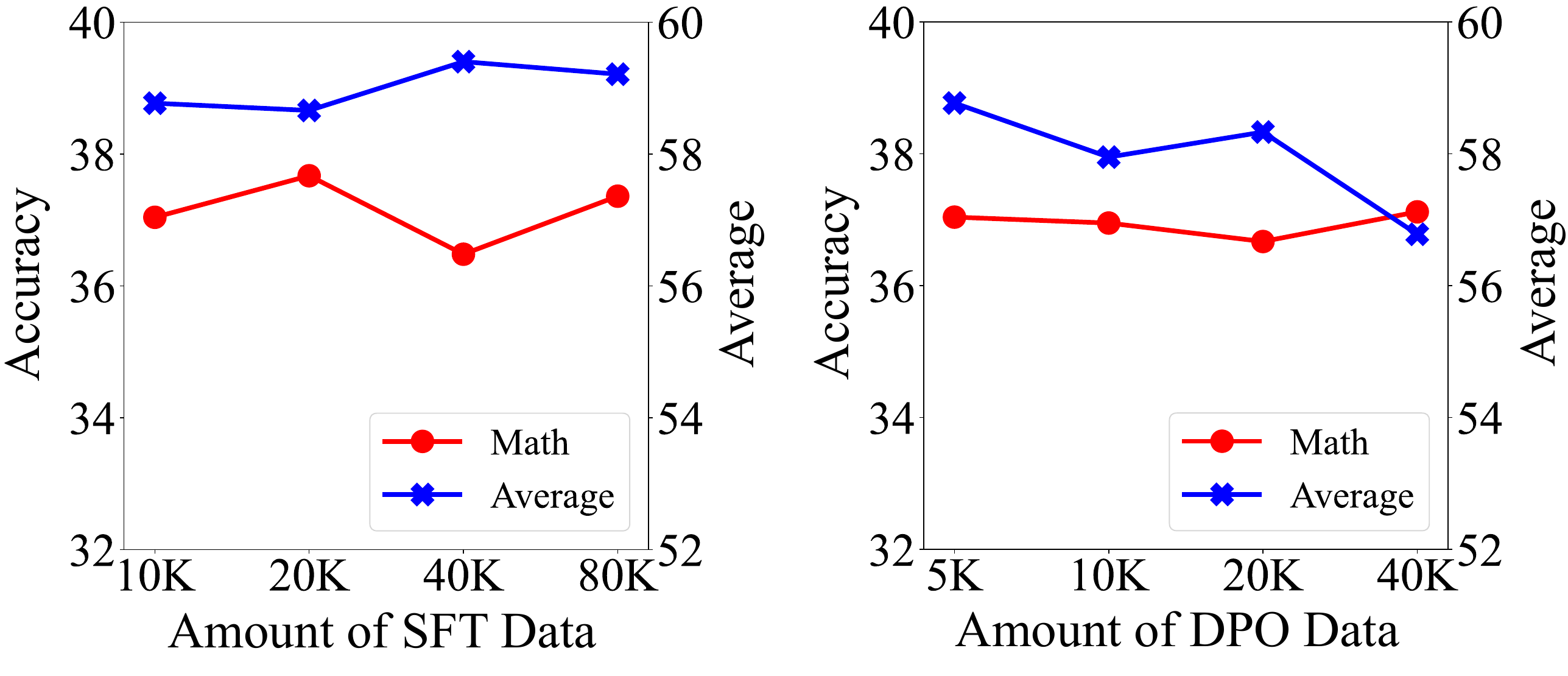}
		\caption{Accuracy on math and Average results \emph{w.r.t.} Amount of SFT data  \textbf{(Left)} and Amount of DPO data \textbf{(Right)}.}
		\label{fig:sft_dpo_amount}
	\end{minipage}
\end{figure}

In this section, we conduct a detailed analysis of the proposed method. Specifically, the investigation focuses on examining the impact of several factors, including the quantity and quality of raw domain data and the selection criteria of samples for format alignment, on the performance of the final chat LLMs. The evaluation employs the same benchmarks as those presented in Table~\ref{tab:chat_res}.

\paratitle{Effect of Quantity and Quality of Raw Domain Data.} We further explore the effect of the quantity and quality of the raw domain documents on the model performance. Considering that the quality of the original StarCoder dataset is approximately consistent, we first utilize it to explore the effect of the amount of domain documents on the domain-specific and general capabilities under similar quality. Moreover, we utilize the AutoMathText dataset to explore the effect of the quality of raw domain data on LLMs' performance by leveraging the annotated quality score for each sample in the AutoMathText dataset.
Specifically, we conduct two group experiments utilizing the StarCoder and AutoMathText datasets as follows:

~$\bullet$ \emph{Proportion of Code Data}: This group aims to compare the variants using different proportions of the original StarCoder corpus, including 10\%, 30\%, and 50\%, while maintaining constancy in other variables. 

~$\bullet$ \emph{Quality of Math Data}: This group aims to compare variants by employing various AutoMathText corpora, each characterized by distinct quality scores with thresholds exceeding 0.5, 0.6, and 0.7 respectively, while ensuring consistency in all other variables.

We show the results in Figure~\ref{fig:code_math_amount}. We can see that increasing the amount of raw domain documents can indeed further enhance the target domain performance under the same quality. However, even though utilizing a larger number of raw domain data (\eg more math texts with $\geq0.5$ score than those with $\geq0.7$ score), the low quality of raw data can also deduce the LLMs' performance regardless of the target domain or general domain, which indicates the quality of domain raw data is a priority over its amount when performing continual pre-training. 

\paratitle{Effect of Format Alignment Data Selection.} Our experiments indicate that selecting 10,000 instruction samples~(about 3\% of the original TULU-V2-mix) and 5,000 alignment samples~(3\% of original UltraFeedback) with the lowest perplexity scores can effectively perform format alignment during supervised fine-tuning and direct preference optimization. Here, we conduct a further ablation study to explore the impact of different selection strategies of format alignment data on the final chat model, which consists of the difficulty and the amount of selected data. Specifically, we conduct three groups of experiments including:

~$\bullet$~\emph{Difficulty of Samples for SFT and DPO}: This group compares four distinct selection strategies: random selection (R), easiest samples with the lowest perplexity (E), hardest samples with the highest perplexity (H), and half easiest samples and half hardest samples (EH).

~$\bullet$ \emph{Amout of Samples for SFT and DPO}: This involves four variants using different quantities of easiest samples from the original TULU-V2-mix dataset (\ie 10K, 20K, 40K, and 80K) and from the original UltraFeedback dataset (\ie 5K, 10K, 20K, and 40K).

~$\bullet$ \emph{Ratios between SFT and DPO Samples}: This experiment utilizes a total of 15,000 samples for format alignment by adopting five different ratios between samples used in supervised fine-tuning and direct preference optimization, ranging from 1:2, 1:1, 2:1, 3:1, to 4:1.

We show the results of each group in Figure~\ref{fig:sft_dpo_ratio}, Figure~\ref{fig:selection_strategy}, and Figure~\ref{fig:sft_dpo_amount}. Firstly, using the easiest samples with the lowest perplexity can balance the domain capability and general capability best compared to other selection strategies. Secondly, it enhances both the domain and general capabilities simultaneously to a certain extent by continuously increasing the amount of SFT training samples. Conversely, this phenomenon is not observed when continuously increasing the amount of DPO training samples, rather, both remain in a state of fluctuation. Finally, we can see that the proportion of SFT and DPO data we selected (\ie 2:1) can optimally balance the general and domain-specific capabilities.
\section{Related Work}

\paratitle{Domain Adaptation of LLMs.} Our work is closely related to efforts in adapting general LLMs to specific domains~\citep{cpt_investigation,KeSL00022,ScialomCM22}. Due to the increasing scale and complexity of LLMs, training domain-specific LLMs from scratch involves significantly high financial and ecological costs~\citep{LuccioniVL23}. To address this issue, recent work has been devoted to studying efficient approaches like continual pre-training, which involves incrementally training general LLMs based on new domain corpora~\citep{que2024d,KeSL00022}, and continual fine-tuning, aiming to fine-tune general LLMs on a series of downstream tasks related to target domains~\citep{RazdaibiedinaMH23,ScialomCM22,abs-2308-08747}. Specially, continual pre-training updates LLMs with large and unlabeled domain-specific corpora, which mainly focuses on memorizing and injecting new knowledge into the parameters of LLMs. However, these approaches might result in catastrophic forgetting and performance degradation in general language tasks~\citep{KarCFMR22,MehtaPCS23}. Another line of work has explored conducting instruction fine-tuning by synthesizing domain-related instructions~\citep{cheng2024instruction,jiang2024instruction}. Nevertheless, these studies require additional models to synthesize amounts of instructions highly related to specific domains, resulting in high computational costs. It is noted that our method differs from these works in several ways. Firstly, we disentangle domain adaptation into knowledge memorization and capability elicitation, focusing on learning domain-specific knowledge and solving domain tasks with learned knowledge, respectively. Secondly, we employ token swap self-distillation in the knowledge mixture pre-training to retain general knowledge and avoid catastrophic forgetting.

\paratitle{Instruction Tuning and Alignment.} Instruction tuning (also known as supervised fine-tuning) employs human-annotated instructions~\citep{SanhWRBSACSRDBX22,MishraKBH22,KopfKRATSBNSNES23,SunSZZCCYG23} or synthetic instructions by proprietary models~\citep{Taori-github-2023-Stanford,vicuna2023,Wang-arxiv-2023-How} to fine-tune LLMs. Besides, alignment with reinforcement learning from human feedback (RLHF)~\citep{Ouyang-arxiv-2022-Training} or direct preference optimization (DPO)~\citep{Rafailov-arxiv-2023-Direct} aims to align LLMs with human preference. Both instruction tuning and alignment are able to elicit knowledge from LLMs and improve their capabilities to solve downstream tasks. Recent work~\citep{LIMA} has demonstrated that LLMs mainly learn the style or format for interacting with users through simple instruction tuning and alignment, by leveraging their prior knowledge and capabilities already acquired during the pre-training stage. Therefore, only employing as few as 1,000 examples in supervised fine-tuning can also achieve satisfactory alignment performance~\citep{LIMA}. Furthermore, by comparing the token distribution before and after alignment, recent work~\citep{abs-2312-01552} found that the most significant distribution shifts appear dominantly in stylistic tokens such as transitional phrases and discourse markers instead of contextual words that involve rich knowledge for solving downstream tasks. Inspired by these studies, we propose to expose knowledge memorization and capability elicitation from instruction tuning and alignment. Unlike these studies which typically focused on instruction tuning or alignment, we differ in that we unify the three stages of training LLMs (\ie continual pre-training, instruction tuning, and alignment) and conduct a knowledge mixture pre-training to mainly focus on learning new domain knowledge while maintaining general knowledge.

\section{Conclusion}
In this study, we refined the conventional approach to domain adaptation for LLMs by introducing a novel two-stage approach, termed Mix-CPT, which encompasses both domain knowledge learning and general format alignment. Distinct from previous strategies, Mix-CPT employed a knowledge mixture of continual pre-training that learns knowledge memorization and utilization simultaneously through the integration of domain-specific raw data with general instruction tuning and alignment data. Besides, we further incorporated the Logit Swap Self-Distillation (LSSD) constraint into the continual pre-training process to relieve catastrophic forgetting. Subsequently, based on the knowledge and capabilities that are already acquired during pre-training, we strategically selected a small number of easy instructions with lower perplexity scores from the instruction set used during the continual pre-training process to make the LLM focus on learning the pure style and format for interacting with humans. We conducted extensive experiments on three benchmark datasets, and the experiment results show that our proposed Mix-CPT outperforms the traditional method, obtaining improvements on both the domain and general capabilities.

\bibliography{iclr2023_conference}

\begin{thebibliography}{62}
\providecommand{\natexlab}[1]{#1}
\providecommand{\url}[1]{\texttt{#1}}
\expandafter\ifx\csname urlstyle\endcsname\relax
  \providecommand{\doi}[1]{doi: #1}\else
  \providecommand{\doi}{doi: \begingroup \urlstyle{rm}\Url}\fi

\bibitem[Austin et~al.(2021)Austin, Odena, Nye, Bosma, Michalewski, Dohan, Jiang, Cai, Terry, Le, and Sutton]{MBPP}
Jacob Austin, Augustus Odena, Maxwell~I. Nye, Maarten Bosma, Henryk Michalewski, David Dohan, Ellen Jiang, Carrie~J. Cai, Michael Terry, Quoc~V. Le, and Charles Sutton.
\newblock Program synthesis with large language models.
\newblock \emph{CoRR}, abs/2108.07732, 2021.

\bibitem[Azerbayev et~al.(2023)Azerbayev, Schoelkopf, Paster, Santos, McAleer, Jiang, Deng, Biderman, and Welleck]{Llemma}
Zhangir Azerbayev, Hailey Schoelkopf, Keiran Paster, Marco~Dos Santos, Stephen McAleer, Albert~Q. Jiang, Jia Deng, Stella Biderman, and Sean Welleck.
\newblock Llemma: An open language model for mathematics.
\newblock \emph{CoRR}, abs/2310.10631, 2023.

\bibitem[Bhakthavatsalam et~al.(2021)Bhakthavatsalam, Khashabi, Khot, Mishra, Richardson, Sabharwal, Schoenick, Tafjord, and Clark]{ARC}
Sumithra Bhakthavatsalam, Daniel Khashabi, Tushar Khot, Bhavana~Dalvi Mishra, Kyle Richardson, Ashish Sabharwal, Carissa Schoenick, Oyvind Tafjord, and Peter Clark.
\newblock Think you have solved direct-answer question answering? try arc-da, the direct-answer {AI2} reasoning challenge.
\newblock \emph{CoRR}, abs/2102.03315, 2021.

\bibitem[Bisk et~al.(2020)Bisk, Zellers, Bras, Gao, and Choi]{PIQA}
Yonatan Bisk, Rowan Zellers, Ronan~Le Bras, Jianfeng Gao, and Yejin Choi.
\newblock {PIQA:} reasoning about physical commonsense in natural language.
\newblock In \emph{The Thirty-Fourth {AAAI} Conference on Artificial Intelligence, {AAAI} 2020, The Thirty-Second Innovative Applications of Artificial Intelligence Conference, {IAAI} 2020, The Tenth {AAAI} Symposium on Educational Advances in Artificial Intelligence, {EAAI} 2020, New York, NY, USA, February 7-12, 2020}, pp.\  7432--7439. {AAAI} Press, 2020.

\bibitem[Brown et~al.(2020)Brown, Mann, Ryder, Subbiah, Kaplan, Dhariwal, Neelakantan, Shyam, Sastry, Askell, Agarwal, Herbert{-}Voss, Krueger, Henighan, Child, Ramesh, Ziegler, Wu, Winter, Hesse, Chen, Sigler, Litwin, Gray, Chess, Clark, Berner, McCandlish, Radford, Sutskever, and Amodei]{gpt3}
Tom~B. Brown, Benjamin Mann, Nick Ryder, Melanie Subbiah, Jared Kaplan, Prafulla Dhariwal, Arvind Neelakantan, Pranav Shyam, Girish Sastry, Amanda Askell, Sandhini Agarwal, Ariel Herbert{-}Voss, Gretchen Krueger, Tom Henighan, Rewon Child, Aditya Ramesh, Daniel~M. Ziegler, Jeffrey Wu, Clemens Winter, Christopher Hesse, Mark Chen, Eric Sigler, Mateusz Litwin, Scott Gray, Benjamin Chess, Jack Clark, Christopher Berner, Sam McCandlish, Alec Radford, Ilya Sutskever, and Dario Amodei.
\newblock Language models are few-shot learners.
\newblock In \emph{Advances in Neural Information Processing Systems 33: Annual Conference on Neural Information Processing Systems 2020, NeurIPS 2020, December 6-12, 2020, virtual}, 2020.

\bibitem[Chen et~al.(2021)Chen, Tworek, Jun, Yuan, de~Oliveira~Pinto, Kaplan, Edwards, Burda, Joseph, Brockman, Ray, Puri, Krueger, Petrov, Khlaaf, Sastry, Mishkin, Chan, Gray, Ryder, Pavlov, Power, Kaiser, Bavarian, Winter, Tillet, Such, Cummings, Plappert, Chantzis, Barnes, Herbert{-}Voss, Guss, Nichol, Paino, Tezak, Tang, Babuschkin, Balaji, Jain, Saunders, Hesse, Carr, Leike, Achiam, Misra, Morikawa, Radford, Knight, Brundage, Murati, Mayer, Welinder, McGrew, Amodei, McCandlish, Sutskever, and Zaremba]{HumanEval}
Mark Chen, Jerry Tworek, Heewoo Jun, Qiming Yuan, Henrique~Pond{\'{e}} de~Oliveira~Pinto, Jared Kaplan, Harrison Edwards, Yuri Burda, Nicholas Joseph, Greg Brockman, Alex Ray, Raul Puri, Gretchen Krueger, Michael Petrov, Heidy Khlaaf, Girish Sastry, Pamela Mishkin, Brooke Chan, Scott Gray, Nick Ryder, Mikhail Pavlov, Alethea Power, Lukasz Kaiser, Mohammad Bavarian, Clemens Winter, Philippe Tillet, Felipe~Petroski Such, Dave Cummings, Matthias Plappert, Fotios Chantzis, Elizabeth Barnes, Ariel Herbert{-}Voss, William~Hebgen Guss, Alex Nichol, Alex Paino, Nikolas Tezak, Jie Tang, Igor Babuschkin, Suchir Balaji, Shantanu Jain, William Saunders, Christopher Hesse, Andrew~N. Carr, Jan Leike, Joshua Achiam, Vedant Misra, Evan Morikawa, Alec Radford, Matthew Knight, Miles Brundage, Mira Murati, Katie Mayer, Peter Welinder, Bob McGrew, Dario Amodei, Sam McCandlish, Ilya Sutskever, and Wojciech Zaremba.
\newblock Evaluating large language models trained on code.
\newblock \emph{CoRR}, abs/2107.03374, 2021.
\newblock URL \url{https://arxiv.org/abs/2107.03374}.

\bibitem[Cheng et~al.(2024)Cheng, Gu, Huang, Bi, Huang, and Wei]{cheng2024instruction}
Daixuan Cheng, Yuxian Gu, Shaohan Huang, Junyu Bi, Minlie Huang, and Furu Wei.
\newblock Instruction pre-training: Language models are supervised multitask learners, 2024.

\bibitem[Chiang et~al.(2023)Chiang, Li, Lin, Sheng, Wu, Zhang, Zheng, Zhuang, Zhuang, Gonzalez, Stoica, and Xing]{vicuna2023}
Wei-Lin Chiang, Zhuohan Li, Zi~Lin, Ying Sheng, Zhanghao Wu, Hao Zhang, Lianmin Zheng, Siyuan Zhuang, Yonghao Zhuang, Joseph~E. Gonzalez, Ion Stoica, and Eric~P. Xing.
\newblock Vicuna: An open-source chatbot impressing gpt-4 with 90\%* chatgpt quality.
\newblock \url{https://vicuna.lmsys.org}, 2023.

\bibitem[Cobbe et~al.(2021)Cobbe, Kosaraju, Bavarian, Chen, Jun, Kaiser, Plappert, Tworek, Hilton, Nakano, Hesse, and Schulman]{GSM8K}
Karl Cobbe, Vineet Kosaraju, Mohammad Bavarian, Mark Chen, Heewoo Jun, Lukasz Kaiser, Matthias Plappert, Jerry Tworek, Jacob Hilton, Reiichiro Nakano, Christopher Hesse, and John Schulman.
\newblock Training verifiers to solve math word problems.
\newblock \emph{CoRR}, abs/2110.14168, 2021.

\bibitem[Contributors(2023)]{OpenCompass}
OpenCompass Contributors.
\newblock Opencompass: A universal evaluation platform for foundation models.
\newblock \url{https://github.com/open-compass/opencompass}, 2023.

\bibitem[Cui et~al.(2023)Cui, Yuan, Ding, Yao, Zhu, Ni, Xie, Liu, and Sun]{UltraFeedback}
Ganqu Cui, Lifan Yuan, Ning Ding, Guanming Yao, Wei Zhu, Yuan Ni, Guotong Xie, Zhiyuan Liu, and Maosong Sun.
\newblock Ultrafeedback: Boosting language models with high-quality feedback.
\newblock \emph{CoRR}, abs/2310.01377, 2023.

\bibitem[Gu et~al.(2023)Gu, Dong, Wei, and Huang]{MiniLM}
Yuxian Gu, Li~Dong, Furu Wei, and Minlie Huang.
\newblock Knowledge distillation of large language models.
\newblock \emph{CoRR}, abs/2306.08543, 2023.

\bibitem[Guo \& Yu(2022)Guo and Yu]{abs-2211-03154}
Xu~Guo and Han Yu.
\newblock On the domain adaptation and generalization of pretrained language models: {A} survey.
\newblock \emph{CoRR}, abs/2211.03154, 2022.

\bibitem[Hendrycks et~al.(2021{\natexlab{a}})Hendrycks, Burns, Basart, Zou, Mazeika, Song, and Steinhardt]{MMLU}
Dan Hendrycks, Collin Burns, Steven Basart, Andy Zou, Mantas Mazeika, Dawn Song, and Jacob Steinhardt.
\newblock Measuring massive multitask language understanding.
\newblock In \emph{9th International Conference on Learning Representations, {ICLR} 2021, Virtual Event, Austria, May 3-7, 2021}. OpenReview.net, 2021{\natexlab{a}}.

\bibitem[Hendrycks et~al.(2021{\natexlab{b}})Hendrycks, Burns, Kadavath, Arora, Basart, Tang, Song, and Steinhardt]{MATH}
Dan Hendrycks, Collin Burns, Saurav Kadavath, Akul Arora, Steven Basart, Eric Tang, Dawn Song, and Jacob Steinhardt.
\newblock Measuring mathematical problem solving with the {MATH} dataset.
\newblock In \emph{Proceedings of the Neural Information Processing Systems Track on Datasets and Benchmarks 1, NeurIPS Datasets and Benchmarks 2021, December 2021, virtual}, 2021{\natexlab{b}}.

\bibitem[Hu et~al.(2024)Hu, Chen, and Ponti]{SIT}
Hanxu Hu, Pinzhen Chen, and Edoardo~M. Ponti.
\newblock Fine-tuning large language models with sequential instructions.
\newblock \emph{CoRR}, abs/2403.07794, 2024.

\bibitem[Huang et~al.(2023)Huang, Bai, Zhu, Zhang, Zhang, Su, Liu, Lv, Zhang, Lei, Fu, Sun, and He]{C-Eval}
Yuzhen Huang, Yuzhuo Bai, Zhihao Zhu, Junlei Zhang, Jinghan Zhang, Tangjun Su, Junteng Liu, Chuancheng Lv, Yikai Zhang, Jiayi Lei, Yao Fu, Maosong Sun, and Junxian He.
\newblock C-eval: {A} multi-level multi-discipline chinese evaluation suite for foundation models.
\newblock In \emph{Advances in Neural Information Processing Systems 36: Annual Conference on Neural Information Processing Systems 2023, NeurIPS 2023, New Orleans, LA, USA, December 10 - 16, 2023}, 2023.

\bibitem[Ivison et~al.(2023)Ivison, Wang, Pyatkin, Lambert, Peters, Dasigi, Jang, Wadden, Smith, Beltagy, and Hajishirzi]{Tulu-V2}
Hamish Ivison, Yizhong Wang, Valentina Pyatkin, Nathan Lambert, Matthew~E. Peters, Pradeep Dasigi, Joel Jang, David Wadden, Noah~A. Smith, Iz~Beltagy, and Hannaneh Hajishirzi.
\newblock Camels in a changing climate: Enhancing {LM} adaptation with tulu 2.
\newblock \emph{CoRR}, abs/2311.10702, 2023.

\bibitem[Jiang et~al.(2024)Jiang, Sun, Shi, Rodriguez, Zhou, Neubig, Lin, Yih, and Iyer]{jiang2024instruction}
Zhengbao Jiang, Zhiqing Sun, Weijia Shi, Pedro Rodriguez, Chunting Zhou, Graham Neubig, Xi~Victoria Lin, Wen-tau Yih, and Srinivasan Iyer.
\newblock Instruction-tuned language models are better knowledge learners.
\newblock \emph{arXiv preprint arXiv:2402.12847}, 2024.

\bibitem[Joshi et~al.(2017)Joshi, Choi, Weld, and Zettlemoyer]{TQ}
Mandar Joshi, Eunsol Choi, Daniel~S. Weld, and Luke Zettlemoyer.
\newblock Triviaqa: {A} large scale distantly supervised challenge dataset for reading comprehension.
\newblock In \emph{Proceedings of the 55th Annual Meeting of the Association for Computational Linguistics, {ACL} 2017, Vancouver, Canada, July 30 - August 4, Volume 1: Long Papers}, pp.\  1601--1611. Association for Computational Linguistics, 2017.

\bibitem[Kar et~al.(2022)Kar, Castellucci, Filice, Malmasi, and Rokhlenko]{KarCFMR22}
Sudipta Kar, Giuseppe Castellucci, Simone Filice, Shervin Malmasi, and Oleg Rokhlenko.
\newblock Preventing catastrophic forgetting in continual learning of new natural language tasks.
\newblock In \emph{{KDD} '22: The 28th {ACM} {SIGKDD} Conference on Knowledge Discovery and Data Mining, Washington, DC, USA, August 14 - 18, 2022}, pp.\  3137--3145. {ACM}, 2022.

\bibitem[Ke et~al.(2022)Ke, Shao, Lin, Xu, Shu, and Liu]{KeSL00022}
Zixuan Ke, Yijia Shao, Haowei Lin, Hu~Xu, Lei Shu, and Bing Liu.
\newblock Adapting a language model while preserving its general knowledge.
\newblock In \emph{Proceedings of the 2022 Conference on Empirical Methods in Natural Language Processing, {EMNLP} 2022, Abu Dhabi, United Arab Emirates, December 7-11, 2022}, pp.\  10177--10188. Association for Computational Linguistics, 2022.

\bibitem[K{\"{o}}pf et~al.(2023)K{\"{o}}pf, Kilcher, von R{\"{u}}tte, Anagnostidis, Tam, Stevens, Barhoum, Nguyen, Stanley, Nagyfi, ES, Suri, Glushkov, Dantuluri, Maguire, Schuhmann, Nguyen, and Mattick]{KopfKRATSBNSNES23}
Andreas K{\"{o}}pf, Yannic Kilcher, Dimitri von R{\"{u}}tte, Sotiris Anagnostidis, Zhi~Rui Tam, Keith Stevens, Abdullah Barhoum, Duc Nguyen, Oliver Stanley, Rich{\'{a}}rd Nagyfi, Shahul ES, Sameer Suri, David Glushkov, Arnav Dantuluri, Andrew Maguire, Christoph Schuhmann, Huu Nguyen, and Alexander Mattick.
\newblock Openassistant conversations - democratizing large language model alignment.
\newblock In \emph{Advances in Neural Information Processing Systems 36: Annual Conference on Neural Information Processing Systems 2023, NeurIPS 2023, New Orleans, LA, USA, December 10 - 16, 2023}, 2023.

\bibitem[Kwiatkowski et~al.(2019)Kwiatkowski, Palomaki, Redfield, Collins, Parikh, Alberti, Epstein, Polosukhin, Devlin, Lee, Toutanova, Jones, Kelcey, Chang, Dai, Uszkoreit, Le, and Petrov]{NQ}
Tom Kwiatkowski, Jennimaria Palomaki, Olivia Redfield, Michael Collins, Ankur~P. Parikh, Chris Alberti, Danielle Epstein, Illia Polosukhin, Jacob Devlin, Kenton Lee, Kristina Toutanova, Llion Jones, Matthew Kelcey, Ming{-}Wei Chang, Andrew~M. Dai, Jakob Uszkoreit, Quoc Le, and Slav Petrov.
\newblock Natural questions: a benchmark for question answering research.
\newblock \emph{Trans. Assoc. Comput. Linguistics}, 7:\penalty0 452--466, 2019.

\bibitem[Lai et~al.(2017)Lai, Xie, Liu, Yang, and Hovy]{RACE}
Guokun Lai, Qizhe Xie, Hanxiao Liu, Yiming Yang, and Eduard~H. Hovy.
\newblock {RACE:} large-scale reading comprehension dataset from examinations.
\newblock In \emph{Proceedings of the 2017 Conference on Empirical Methods in Natural Language Processing, {EMNLP} 2017, Copenhagen, Denmark, September 9-11, 2017}, pp.\  785--794. Association for Computational Linguistics, 2017.

\bibitem[Li et~al.(2023)Li, Allal, Zi, Muennighoff, Kocetkov, Mou, Marone, Akiki, Li, Chim, Liu, Zheltonozhskii, Zhuo, Wang, Dehaene, Davaadorj, Lamy{-}Poirier, Monteiro, Shliazhko, Gontier, Meade, Zebaze, Yee, Umapathi, Zhu, Lipkin, Oblokulov, Wang, V, Stillerman, Patel, Abulkhanov, Zocca, Dey, Zhang, Moustafa{-}Fahmy, Bhattacharyya, Yu, Singh, Luccioni, Villegas, Kunakov, Zhdanov, Romero, Lee, Timor, Ding, Schlesinger, Schoelkopf, Ebert, Dao, Mishra, Gu, Robinson, Anderson, Dolan{-}Gavitt, Contractor, Reddy, Fried, Bahdanau, Jernite, Ferrandis, Hughes, Wolf, Guha, von Werra, and de~Vries]{StarCoder}
Raymond Li, Loubna~Ben Allal, Yangtian Zi, Niklas Muennighoff, Denis Kocetkov, Chenghao Mou, Marc Marone, Christopher Akiki, Jia Li, Jenny Chim, Qian Liu, Evgenii Zheltonozhskii, Terry~Yue Zhuo, Thomas Wang, Olivier Dehaene, Mishig Davaadorj, Joel Lamy{-}Poirier, Jo{\~{a}}o Monteiro, Oleh Shliazhko, Nicolas Gontier, Nicholas Meade, Armel Zebaze, Ming{-}Ho Yee, Logesh~Kumar Umapathi, Jian Zhu, Benjamin Lipkin, Muhtasham Oblokulov, Zhiruo Wang, Rudra~Murthy V, Jason Stillerman, Siva~Sankalp Patel, Dmitry Abulkhanov, Marco Zocca, Manan Dey, Zhihan Zhang, Nour Moustafa{-}Fahmy, Urvashi Bhattacharyya, Wenhao Yu, Swayam Singh, Sasha Luccioni, Paulo Villegas, Maxim Kunakov, Fedor Zhdanov, Manuel Romero, Tony Lee, Nadav Timor, Jennifer Ding, Claire Schlesinger, Hailey Schoelkopf, Jan Ebert, Tri Dao, Mayank Mishra, Alex Gu, Jennifer Robinson, Carolyn~Jane Anderson, Brendan Dolan{-}Gavitt, Danish Contractor, Siva Reddy, Daniel Fried, Dzmitry Bahdanau, Yacine Jernite, Carlos~Mu{\~{n}}oz Ferrandis, Sean Hughes, Thomas
  Wolf, Arjun Guha, Leandro von Werra, and Harm de~Vries.
\newblock Starcoder: may the source be with you!
\newblock \emph{CoRR}, abs/2305.06161, 2023.

\bibitem[Lin et~al.(2023)Lin, Ravichander, Lu, Dziri, Sclar, Chandu, Bhagavatula, and Choi]{abs-2312-01552}
Bill~Yuchen Lin, Abhilasha Ravichander, Ximing Lu, Nouha Dziri, Melanie Sclar, Khyathi Chandu, Chandra Bhagavatula, and Yejin Choi.
\newblock The unlocking spell on base llms: Rethinking alignment via in-context learning.
\newblock \emph{CoRR}, abs/2312.01552, 2023.
\newblock \doi{10.48550/ARXIV.2312.01552}.
\newblock URL \url{https://doi.org/10.48550/arXiv.2312.01552}.

\bibitem[Lin et~al.(2024)Lin, Gou, Gong, Liu, Shen, Xu, Lin, Yang, Jiao, Duan, and Chen]{Rho}
Zhenghao Lin, Zhibin Gou, Yeyun Gong, Xiao Liu, Yelong Shen, Ruochen Xu, Chen Lin, Yujiu Yang, Jian Jiao, Nan Duan, and Weizhu Chen.
\newblock Rho-1: Not all tokens are what you need.
\newblock \emph{CoRR}, abs/2404.07965, 2024.

\bibitem[Luccioni et~al.(2023)Luccioni, Viguier, and Ligozat]{LuccioniVL23}
Alexandra~Sasha Luccioni, Sylvain Viguier, and Anne{-}Laure Ligozat.
\newblock Estimating the carbon footprint of bloom, a 176b parameter language model.
\newblock \emph{J. Mach. Learn. Res.}, 24:\penalty0 253:1--253:15, 2023.

\bibitem[Luo et~al.(2023{\natexlab{a}})Luo, Yang, Meng, Li, Zhou, and Zhang]{abs-2308-08747}
Yun Luo, Zhen Yang, Fandong Meng, Yafu Li, Jie Zhou, and Yue Zhang.
\newblock An empirical study of catastrophic forgetting in large language models during continual fine-tuning.
\newblock \emph{CoRR}, abs/2308.08747, 2023{\natexlab{a}}.

\bibitem[Luo et~al.(2023{\natexlab{b}})Luo, Xu, Zhao, Sun, Geng, Hu, Tao, Ma, Lin, and Jiang]{WizardCoder}
Ziyang Luo, Can Xu, Pu~Zhao, Qingfeng Sun, Xiubo Geng, Wenxiang Hu, Chongyang Tao, Jing Ma, Qingwei Lin, and Daxin Jiang.
\newblock Wizardcoder: Empowering code large language models with evol-instruct.
\newblock \emph{CoRR}, abs/2306.08568, 2023{\natexlab{b}}.

\bibitem[Mehta et~al.(2023)Mehta, Patil, Chandar, and Strubell]{MehtaPCS23}
Sanket~Vaibhav Mehta, Darshan Patil, Sarath Chandar, and Emma Strubell.
\newblock An empirical investigation of the role of pre-training in lifelong learning.
\newblock \emph{J. Mach. Learn. Res.}, 24:\penalty0 214:1--214:50, 2023.

\bibitem[Meng et~al.(2024)Meng, Xia, and Chen]{SimPO}
Yu~Meng, Mengzhou Xia, and Danqi Chen.
\newblock Simpo: Simple preference optimization with a reference-free reward.
\newblock \emph{CoRR}, abs/2405.14734, 2024.

\bibitem[Mihaylov et~al.(2018)Mihaylov, Clark, Khot, and Sabharwal]{OpenBookQA}
Todor Mihaylov, Peter Clark, Tushar Khot, and Ashish Sabharwal.
\newblock Can a suit of armor conduct electricity? {A} new dataset for open book question answering.
\newblock In \emph{Proceedings of the 2018 Conference on Empirical Methods in Natural Language Processing, Brussels, Belgium, October 31 - November 4, 2018}, pp.\  2381--2391. Association for Computational Linguistics, 2018.

\bibitem[Mishra et~al.(2022)Mishra, Khashabi, Baral, and Hajishirzi]{MishraKBH22}
Swaroop Mishra, Daniel Khashabi, Chitta Baral, and Hannaneh Hajishirzi.
\newblock Cross-task generalization via natural language crowdsourcing instructions.
\newblock In Smaranda Muresan, Preslav Nakov, and Aline Villavicencio (eds.), \emph{Proceedings of the 60th Annual Meeting of the Association for Computational Linguistics (Volume 1: Long Papers), {ACL} 2022, Dublin, Ireland, May 22-27, 2022}, pp.\  3470--3487. Association for Computational Linguistics, 2022.

\bibitem[OpenAI(2023)]{OpenAI-OpenAI-2023-GPT-4}
OpenAI.
\newblock Gpt-4 technical report.
\newblock \emph{OpenAI Blog}, 2023.

\bibitem[Ouyang et~al.(2022{\natexlab{a}})Ouyang, Wu, Jiang, Almeida, Wainwright, Mishkin, Zhang, Agarwal, Slama, Ray, Schulman, Hilton, Kelton, Miller, Simens, Askell, Welinder, Christiano, Leike, and Lowe]{Ouyang-arxiv-2022-Training}
Long Ouyang, Jeff Wu, Xu~Jiang, Diogo Almeida, Carroll~L. Wainwright, Pamela Mishkin, Chong Zhang, Sandhini Agarwal, Katarina Slama, Alex Ray, John Schulman, Jacob Hilton, Fraser Kelton, Luke Miller, Maddie Simens, Amanda Askell, Peter Welinder, Paul~F. Christiano, Jan Leike, and Ryan Lowe.
\newblock Training language models to follow instructions with human feedback.
\newblock \emph{arXiv preprint arXiv:2203.02155}, 2022{\natexlab{a}}.

\bibitem[Ouyang et~al.(2022{\natexlab{b}})Ouyang, Wu, Jiang, Almeida, Wainwright, Mishkin, Zhang, Agarwal, Slama, Ray, Schulman, Hilton, Kelton, Miller, Simens, Askell, Welinder, Christiano, Leike, and Lowe]{InstructGPT}
Long Ouyang, Jeffrey Wu, Xu~Jiang, Diogo Almeida, Carroll~L. Wainwright, Pamela Mishkin, Chong Zhang, Sandhini Agarwal, Katarina Slama, Alex Ray, John Schulman, Jacob Hilton, Fraser Kelton, Luke Miller, Maddie Simens, Amanda Askell, Peter Welinder, Paul~F. Christiano, Jan Leike, and Ryan Lowe.
\newblock Training language models to follow instructions with human feedback.
\newblock In \emph{Advances in Neural Information Processing Systems 35: Annual Conference on Neural Information Processing Systems 2022, NeurIPS 2022, New Orleans, LA, USA, November 28 - December 9, 2022}, 2022{\natexlab{b}}.

\bibitem[Que et~al.(2024)Que, Liu, Zhang, Zhang, Qu, Ma, Duan, Bai, Wang, Zhang, et~al.]{que2024d}
Haoran Que, Jiaheng Liu, Ge~Zhang, Chenchen Zhang, Xingwei Qu, Yinghao Ma, Feiyu Duan, Zhiqi Bai, Jiakai Wang, Yuanxing Zhang, et~al.
\newblock D-cpt law: Domain-specific continual pre-training scaling law for large language models.
\newblock \emph{arXiv preprint arXiv:2406.01375}, 2024.

\bibitem[Rafailov et~al.(2023{\natexlab{a}})Rafailov, Sharma, Mitchell, Ermon, Manning, and Finn]{Rafailov-arxiv-2023-Direct}
Rafael Rafailov, Archit Sharma, Eric Mitchell, Stefano Ermon, Christopher~D. Manning, and Chelsea Finn.
\newblock Direct preference optimization: Your language model is secretly a reward model.
\newblock \emph{arXiv preprint arXiv:2305.18290}, 2023{\natexlab{a}}.

\bibitem[Rafailov et~al.(2023{\natexlab{b}})Rafailov, Sharma, Mitchell, Manning, Ermon, and Finn]{DPO}
Rafael Rafailov, Archit Sharma, Eric Mitchell, Christopher~D. Manning, Stefano Ermon, and Chelsea Finn.
\newblock Direct preference optimization: Your language model is secretly a reward model.
\newblock In \emph{Advances in Neural Information Processing Systems 36: Annual Conference on Neural Information Processing Systems 2023, NeurIPS 2023, New Orleans, LA, USA, December 10 - 16, 2023}, 2023{\natexlab{b}}.

\bibitem[Razdaibiedina et~al.(2023)Razdaibiedina, Mao, Hou, Khabsa, Lewis, and Almahairi]{RazdaibiedinaMH23}
Anastasia Razdaibiedina, Yuning Mao, Rui Hou, Madian Khabsa, Mike Lewis, and Amjad Almahairi.
\newblock Progressive prompts: Continual learning for language models.
\newblock In \emph{The Eleventh International Conference on Learning Representations, {ICLR} 2023, Kigali, Rwanda, May 1-5, 2023}. OpenReview.net, 2023.

\bibitem[Ren et~al.(2024)Ren, Cao, Lin, Liu, Han, Zeng, Wan, Cai, and Sun]{Ren-2024-learning}
Mengjie Ren, Boxi Cao, Hongyu Lin, Cao Liu, Xianpei Han, Ke~Zeng, Guanglu Wan, Xunliang Cai, and Le~Sun.
\newblock Learning or self-aligning? rethinking instruction fine-tuning.
\newblock \emph{CoRR}, abs/2402.18243, 2024.

\bibitem[Rozi{\`{e}}re et~al.(2023)Rozi{\`{e}}re, Gehring, Gloeckle, Sootla, Gat, Tan, Adi, Liu, Remez, Rapin, Kozhevnikov, Evtimov, Bitton, Bhatt, Canton{-}Ferrer, Grattafiori, Xiong, D{\'{e}}fossez, Copet, Azhar, Touvron, Martin, Usunier, Scialom, and Synnaeve]{CodeLLaMA}
Baptiste Rozi{\`{e}}re, Jonas Gehring, Fabian Gloeckle, Sten Sootla, Itai Gat, Xiaoqing~Ellen Tan, Yossi Adi, Jingyu Liu, Tal Remez, J{\'{e}}r{\'{e}}my Rapin, Artyom Kozhevnikov, Ivan Evtimov, Joanna Bitton, Manish Bhatt, Cristian Canton{-}Ferrer, Aaron Grattafiori, Wenhan Xiong, Alexandre D{\'{e}}fossez, Jade Copet, Faisal Azhar, Hugo Touvron, Louis Martin, Nicolas Usunier, Thomas Scialom, and Gabriel Synnaeve.
\newblock Code llama: Open foundation models for code.
\newblock \emph{CoRR}, abs/2308.12950, 2023.

\bibitem[Sanh et~al.(2022)Sanh, Webson, Raffel, Bach, Sutawika, Alyafeai, Chaffin, Stiegler, Raja, Dey, Bari, Xu, Thakker, Sharma, Szczechla, Kim, Chhablani, Nayak, Datta, Chang, Jiang, Wang, Manica, Shen, Yong, Pandey, Bawden, Wang, Neeraj, Rozen, Sharma, Santilli, F{\'{e}}vry, Fries, Teehan, Scao, Biderman, Gao, Wolf, and Rush]{SanhWRBSACSRDBX22}
Victor Sanh, Albert Webson, Colin Raffel, Stephen~H. Bach, Lintang Sutawika, Zaid Alyafeai, Antoine Chaffin, Arnaud Stiegler, Arun Raja, Manan Dey, M~Saiful Bari, Canwen Xu, Urmish Thakker, Shanya~Sharma Sharma, Eliza Szczechla, Taewoon Kim, Gunjan Chhablani, Nihal~V. Nayak, Debajyoti Datta, Jonathan Chang, Mike~Tian{-}Jian Jiang, Han Wang, Matteo Manica, Sheng Shen, Zheng~Xin Yong, Harshit Pandey, Rachel Bawden, Thomas Wang, Trishala Neeraj, Jos Rozen, Abheesht Sharma, Andrea Santilli, Thibault F{\'{e}}vry, Jason~Alan Fries, Ryan Teehan, Teven~Le Scao, Stella Biderman, Leo Gao, Thomas Wolf, and Alexander~M. Rush.
\newblock Multitask prompted training enables zero-shot task generalization.
\newblock In \emph{The Tenth International Conference on Learning Representations, {ICLR} 2022, Virtual Event, April 25-29, 2022}. OpenReview.net, 2022.

\bibitem[Scialom et~al.(2022)Scialom, Chakrabarty, and Muresan]{ScialomCM22}
Thomas Scialom, Tuhin Chakrabarty, and Smaranda Muresan.
\newblock Fine-tuned language models are continual learners.
\newblock In \emph{Proceedings of the 2022 Conference on Empirical Methods in Natural Language Processing, {EMNLP} 2022, Abu Dhabi, United Arab Emirates, December 7-11, 2022}, pp.\  6107--6122. Association for Computational Linguistics, 2022.

\bibitem[Sun et~al.(2023)Sun, Shen, Zhou, Zhang, Chen, Cox, Yang, and Gan]{SunSZZCCYG23}
Zhiqing Sun, Yikang Shen, Qinhong Zhou, Hongxin Zhang, Zhenfang Chen, David~D. Cox, Yiming Yang, and Chuang Gan.
\newblock Principle-driven self-alignment of language models from scratch with minimal human supervision.
\newblock In \emph{Advances in Neural Information Processing Systems 36: Annual Conference on Neural Information Processing Systems 2023, NeurIPS 2023, New Orleans, LA, USA, December 10 - 16, 2023}, 2023.

\bibitem[Suzgun et~al.(2023)Suzgun, Scales, Sch{\"{a}}rli, Gehrmann, Tay, Chung, Chowdhery, Le, Chi, Zhou, and Wei]{BBH}
Mirac Suzgun, Nathan Scales, Nathanael Sch{\"{a}}rli, Sebastian Gehrmann, Yi~Tay, Hyung~Won Chung, Aakanksha Chowdhery, Quoc~V. Le, Ed~H. Chi, Denny Zhou, and Jason Wei.
\newblock Challenging big-bench tasks and whether chain-of-thought can solve them.
\newblock In \emph{Findings of the Association for Computational Linguistics: {ACL} 2023, Toronto, Canada, July 9-14, 2023}, pp.\  13003--13051. Association for Computational Linguistics, 2023.

\bibitem[Talmor et~al.(2019)Talmor, Herzig, Lourie, and Berant]{CSQA}
Alon Talmor, Jonathan Herzig, Nicholas Lourie, and Jonathan Berant.
\newblock Commonsenseqa: {A} question answering challenge targeting commonsense knowledge.
\newblock In \emph{Proceedings of the 2019 Conference of the North American Chapter of the Association for Computational Linguistics: Human Language Technologies, {NAACL-HLT} 2019, Minneapolis, MN, USA, June 2-7, 2019, Volume 1 (Long and Short Papers)}, pp.\  4149--4158. Association for Computational Linguistics, 2019.

\bibitem[Taori et~al.(2023)Taori, Gulrajani, Zhang, Dubois, Li, Guestrin, Liang, and Hashimoto]{Taori-github-2023-Stanford}
Rohan Taori, Ishaan Gulrajani, Tianyi Zhang, Yann Dubois, Xuechen Li, Carlos Guestrin, Percy Liang, and Tatsunori~B. Hashimoto.
\newblock Stanford alpaca: An instruction-following llama model.
\newblock \url{https://github.com/tatsu-lab/stanford_alpaca}, 2023.

\bibitem[Touvron et~al.(2023)Touvron, Lavril, Izacard, Martinet, Lachaux, Lacroix, Rozi{\`{e}}re, Goyal, Hambro, Azhar, Rodriguez, Joulin, Grave, and Lample]{LLaMA}
Hugo Touvron, Thibaut Lavril, Gautier Izacard, Xavier Martinet, Marie{-}Anne Lachaux, Timoth{\'{e}}e Lacroix, Baptiste Rozi{\`{e}}re, Naman Goyal, Eric Hambro, Faisal Azhar, Aur{\'{e}}lien Rodriguez, Armand Joulin, Edouard Grave, and Guillaume Lample.
\newblock Llama: Open and efficient foundation language models.
\newblock \emph{CoRR}, abs/2302.13971, 2023.

\bibitem[Wang et~al.(2024)Wang, Mi, Chen, Xue, Wang, Zhu, Wong, and Xu]{abs-2403-02756}
Rui Wang, Fei Mi, Yi~Chen, Boyang Xue, Hongru Wang, Qi~Zhu, Kam{-}Fai Wong, and Ruifeng Xu.
\newblock Role prompting guided domain adaptation with general capability preserve for large language models.
\newblock \emph{CoRR}, abs/2403.02756, 2024.

\bibitem[Wang et~al.(2023{\natexlab{a}})Wang, Wei, Schuurmans, Le, Chi, Narang, Chowdhery, and Zhou]{self-consistency}
Xuezhi Wang, Jason Wei, Dale Schuurmans, Quoc~V. Le, Ed~H. Chi, Sharan Narang, Aakanksha Chowdhery, and Denny Zhou.
\newblock Self-consistency improves chain of thought reasoning in language models.
\newblock In \emph{The Eleventh International Conference on Learning Representations, {ICLR} 2023, Kigali, Rwanda, May 1-5, 2023}. OpenReview.net, 2023{\natexlab{a}}.

\bibitem[Wang et~al.(2023{\natexlab{b}})Wang, Ivison, Dasigi, Hessel, Khot, Chandu, Wadden, MacMillan, Smith, Beltagy, and Hajishirzi]{Wang-arxiv-2023-How}
Yizhong Wang, Hamish Ivison, Pradeep Dasigi, Jack Hessel, Tushar Khot, Khyathi~Raghavi Chandu, David Wadden, Kelsey MacMillan, Noah~A. Smith, Iz~Beltagy, and Hannaneh Hajishirzi.
\newblock How far can camels go? exploring the state of instruction tuning on open resources.
\newblock \emph{arXiv preprint arXiv:2306.04751}, 2023{\natexlab{b}}.

\bibitem[Wei et~al.(2022)Wei, Wang, Schuurmans, Bosma, Ichter, Xia, Chi, Le, and Zhou]{cot}
Jason Wei, Xuezhi Wang, Dale Schuurmans, Maarten Bosma, Brian Ichter, Fei Xia, Ed~H. Chi, Quoc~V. Le, and Denny Zhou.
\newblock Chain-of-thought prompting elicits reasoning in large language models.
\newblock In \emph{Advances in Neural Information Processing Systems 35: Annual Conference on Neural Information Processing Systems 2022, NeurIPS 2022, New Orleans, LA, USA, November 28 - December 9, 2022}, 2022.

\bibitem[Yildiz et~al.(2024)Yildiz, Ravichandran, Punia, Bethge, and Ermis]{cpt_investigation}
{\c{C}}agatay Yildiz, Nishaanth~Kanna Ravichandran, Prishruit Punia, Matthias Bethge, and Beyza Ermis.
\newblock Investigating continual pretraining in large language models: Insights and implications.
\newblock \emph{CoRR}, abs/2402.17400, 2024.

\bibitem[Zellers et~al.(2019)Zellers, Holtzman, Bisk, Farhadi, and Choi]{HellaSwag}
Rowan Zellers, Ari Holtzman, Yonatan Bisk, Ali Farhadi, and Yejin Choi.
\newblock Hellaswag: Can a machine really finish your sentence?
\newblock In \emph{Proceedings of the 57th Conference of the Association for Computational Linguistics, {ACL} 2019, Florence, Italy, July 28- August 2, 2019, Volume 1: Long Papers}, pp.\  4791--4800. Association for Computational Linguistics, 2019.

\bibitem[Zhang et~al.(2024)Zhang, Luo, Yuan, and Yao]{AutoMathText}
Yifan Zhang, Yifan Luo, Yang Yuan, and Andrew~Chi{-}Chih Yao.
\newblock Automathtext: Autonomous data selection with language models for mathematical texts.
\newblock \emph{CoRR}, abs/2402.07625, 2024.

\bibitem[Zhao et~al.(2023)Zhao, Zhou, Li, Tang, Wang, Hou, Min, Zhang, Zhang, Dong, Du, Yang, Chen, Chen, Jiang, Ren, Li, Tang, Liu, Liu, Nie, and Wen]{llm_survey}
Wayne~Xin Zhao, Kun Zhou, Junyi Li, Tianyi Tang, Xiaolei Wang, Yupeng Hou, Yingqian Min, Beichen Zhang, Junjie Zhang, Zican Dong, Yifan Du, Chen Yang, Yushuo Chen, Zhipeng Chen, Jinhao Jiang, Ruiyang Ren, Yifan Li, Xinyu Tang, Zikang Liu, Peiyu Liu, Jian{-}Yun Nie, and Ji{-}Rong Wen.
\newblock A survey of large language models.
\newblock \emph{CoRR}, abs/2303.18223, 2023.

\bibitem[Zheng et~al.(2023)Zheng, Chiang, Sheng, Zhuang, Wu, Zhuang, Lin, Li, Li, Xing, Zhang, Gonzalez, and Stoica]{MT-Bench}
Lianmin Zheng, Wei{-}Lin Chiang, Ying Sheng, Siyuan Zhuang, Zhanghao Wu, Yonghao Zhuang, Zi~Lin, Zhuohan Li, Dacheng Li, Eric~P. Xing, Hao Zhang, Joseph~E. Gonzalez, and Ion Stoica.
\newblock Judging llm-as-a-judge with mt-bench and chatbot arena.
\newblock In \emph{Advances in Neural Information Processing Systems 36: Annual Conference on Neural Information Processing Systems 2023, NeurIPS 2023, New Orleans, LA, USA, December 10 - 16, 2023}, 2023.

\bibitem[Zhou et~al.(2023)Zhou, Liu, Xu, Iyer, Sun, Mao, Ma, Efrat, Yu, Yu, Zhang, Ghosh, Lewis, Zettlemoyer, and Levy]{LIMA}
Chunting Zhou, Pengfei Liu, Puxin Xu, Srinivasan Iyer, Jiao Sun, Yuning Mao, Xuezhe Ma, Avia Efrat, Ping Yu, Lili Yu, Susan Zhang, Gargi Ghosh, Mike Lewis, Luke Zettlemoyer, and Omer Levy.
\newblock {LIMA:} less is more for alignment.
\newblock In \emph{Advances in Neural Information Processing Systems 36: Annual Conference on Neural Information Processing Systems 2023, NeurIPS 2023, New Orleans, LA, USA, December 10 - 16, 2023}, 2023.

\bibitem[Zhou et~al.(2024)Zhou, Zhang, Wang, Chen, Zhao, Sha, Sheng, Wang, and Wen]{JiuZhang3.0}
Kun Zhou, Beichen Zhang, Jiapeng Wang, Zhipeng Chen, Wayne~Xin Zhao, Jing Sha, Zhichao Sheng, Shijin Wang, and Ji{-}Rong Wen.
\newblock Jiuzhang3.0: Efficiently improving mathematical reasoning by training small data synthesis models.
\newblock \emph{CoRR}, abs/2405.14365, 2024.

\end{thebibliography}
\bibliographystyle{iclr2023_conference}

\end{document}